\pdfoutput=1

\documentclass[11pt]{article}

\usepackage[]{coling}

\usepackage{times}
\usepackage{latexsym}
\usepackage{makecell}
\usepackage{amsmath}
\usepackage[T1]{fontenc}

\usepackage[utf8]{inputenc}

\usepackage{hyperref}
\usepackage{booktabs}
\usepackage{graphicx}
\usepackage{footmisc}
\usepackage{subcaption,siunitx,booktabs}

\usepackage[nameinlink]{cleveref}
\Crefformat{section}{\S#2#1#3} 
\Crefname{algorithm}{Alg.}{Algs.}
\Crefformat{subsection}{\S#2#1#3}
\Crefname{equation}{Eq.}{Eqs.}
\Crefname{figure}{Fig.}{Figs.}
\Crefname{table}{Tab.}{Tab.}
\Crefname{theorem}{Theorem}{Theorem}
\Crefformat{theorem}{Thm. #2#1#3}
\Crefformat{assum}{Asm. #2#1#3}
\Crefformat{appendix}{Appx. #2#1#3}
\Crefformat{hypothesis}{Hyp. #2#1#3}
\Crefformat{subsection}{\S#2#1#3}
\Crefname{equation}{Eq.}{Eqs.}
\Crefname{figure}{Fig.}{Figs.}
\usepackage[colorinlistoftodos,prependcaption,textsize=tiny]{todonotes}

\usepackage{caption}
\usepackage{soul}

\newcommand{\llong}[1]{\textcolor{black}{#1}}

\usepackage{float}
\usepackage{array}

\usepackage{multirow}
\usepackage{hhline}
\usepackage{microtype}
\usepackage{tcolorbox}
\usepackage{arydshln}
\usepackage{amssymb}
\usepackage{inconsolata}

\definecolor{inc}{RGB}{84,123,71}
\definecolor{dec}{RGB}{219, 48, 122}

\newcommand{\model}{COO}

\title{Aligning Large Language Models with Human Opinions through \\Persona Selection and Value--Belief--Norm Reasoning}

\author{
Do Xuan Long$^{1,2}$, Kenji Kawaguchi$^{1}$, Min-Yen Kan$^{1}$, Nancy F. Chen$^{2}$ \\
$^{1}$National University of Singapore, \\ $^{2}$Institute for Infocomm Research (I$^2$R), A*STAR \\
\texttt{xuanlong.do@u.nus.edu, \{kenji,knmnyn\}@nus.edu.sg,}\\
\texttt{nfychen@i2r.a-star.edu.sg}
}

\begin{document}
\maketitle
\begin{abstract}
Reasoning and predicting human opinions with large language models (LLMs) is essential yet challenging. Current methods employ role-playing with personae but face two major issues: LLMs are sensitive to even a single irrelevant persona, \llong{changing up to 30\% of the predictions}; and LLMs fail to reason strategically over personae. We propose Chain-of-Opinion (\model{}\footnote{Our codes and data will be made publicly available \href{https://github.com/dxlong2000/COO}{here}.}), a simple four-step solution modeling \emph{which} and \emph{how} to reason with personae, inspired by the Value--Belief--Norm (VBN) theory. \model{} differentiates between \emph{explicit personae} (demographics and ideology) and \emph{implicit personae} (historical opinions), involves: (1) filtering irrelevant attributes from explicit personae; (2) ranking implicit personae into a preferential list for selecting top-k; (3) applying novel VBN reasoning to extract user environmental and personal value, belief, and norm variables for accurate and reliable predictions; and (4) iterating VBN reasoning with progressively larger lists of implicit personae to handle potential persona insufficiency. \model{} efficiently achieves new state-of-the-art opinion prediction via prompting with only 5 inference calls, improving prior techniques by up to 4\%. Notably, fine-tuning LMs with \model{}'s data results in significantly better opinion-aligned models, by up to 23\%.

\end{abstract}

\section{Introduction} \label{sec:intro}




Pre-trained large language models (LLMs) are becoming indispensable tools, serving in various assistant roles such as dialogue agents \cite{openai2022chatgpt,google2022bard} and data analysts \cite{cheng-etal-2023-gpt}. Notably, they demonstrate the capability to model distinct opinions that influence response generation  \cite{bai2022constitutional,glaese2022improving,santurkar2023whose}. Unfortunately, their opinions are shaped by extensive training data, which are themselves influenced by countless human perspectives and thus challenging to comprehend.
As human--AI interactions grow, instructing models to reason in alignment with human opinions is crucial for effective personalization.

Although fine-tuning alignment methods such as RLHF \cite{christiano2017deep,ouyang2022training} are widely employed, their application to personalized opinions remains challenging due to the significant compute and data required. Prompt-based role-playing frameworks using \emph{personae} have emerged as alternatives. Early work in this direction focused on aligning models with social groups rather than individuals \cite{perez-etal-2023-discovering,santurkar2023whose}. However, \citet{santurkar2023whose} found that simple persona-based prompting exhibits low steerability, even for well-represented groups. Further, \citet{hwang2023aligning} revealed significant opinion variation among individuals with similar demographics, underscoring the challenge of aligning LLMs to individual opinions. They subsequently proposed a na\"{\i}ve solution to model individuals by incorporating the user's demographics and ideology (we term these \textbf{explicit personae}), alongside their historical opinions (\textbf{implicit personae}).


While na\"{\i}vely incorporating explicit and implicit personae into the prompt shows promise for individualization \cite{santurkar2023whose,hwang2023aligning}, this approach is suboptimal. Both persona types include ones that may be irrelevant to the opinion of interest, such as the \texttt{``Citizenship''} for the \texttt{``Gun''} question in \Cref{fig:model}. This poses a problem as we observe that LLMs are highly sensitive to such noise (detailed in \Cref{sec:llms-distracted}), highlighting the task of \textbf{relevant persona selection} as important yet unsolved. Moreover, \textbf{effective reasoning} over explicit and implicit personae for opinion prediction is challenging: we find that Chain-of-Thought \cite{wei2022chain,kojima2022large} unexpectedly fails to improve opinion prediction in LLMs, due to reasoning inconsistencies. 


In line with the Value-Belief-Norm (VBN) theory \cite{stern1999value} which asserts that values, beliefs, and norms influence human behavior distinctly, {we argue that the explicit and implicit personae should be processed and utilized differently}. The explicit offers clear insights into user \emph{environmental values} (binary relevancy), while the implicit reveals \emph{nuanced and context-specific beliefs and norms} shaped by personal experiences (complex relevancy). Building on this, we introduce \textbf{Chain-of-Opinion} (\model{}, \Cref{fig:model}), a novel four-step framework that optionally leverages both persona types to address the above challenges: (1) an LLM analyzes explicit personae to discard irrelevant ones; (2) the LLM ranks implicit persona opinions by usefulness and selects the top-$K$; (3) VBN reasoning where the LLM generates high-level environmental values from selected explicit personae and nuanced individual beliefs and norms from top-K implicit personae before deriving a prediction; and (4) \model{} applies VBN reasoning using varying $K$ of implicit personae from step (2) to prevent the model from refusing to answer due to insufficient personae.


\model{}
achieves state-of-the-art opinion prediction with persona-based prompting by just five inference calls (\Cref{appendix:prompting-cost}). Moreover, fine-tuning with data from \model{}'s steps (1--3) enhances LMs by up to 23\%, resulting in Flan-T5 base model \cite{chung2022scaling} comparable with GPT-4 \cite{openai2023gpt4}. \model{} is highly generalizable in scenarios with missing personae (\Cref{ssec:coo-generalization}), and its four steps can motivate and be applicable to other personalized tasks involving explicit personae and user historical views (\Cref{appdx:generalization-to-other-tasks}).

\section{Related Work} \label{sec:related-work}

\paragraph{LLM role-play with personae.} 
Aligning language models with human behavior via personae is a growing study area.  Such alignment increases user satisfaction and personalization \cite{wang2023aligning,chen2024persona}. 
One line of work develops prompting techniques with user demographics, encouraging LLMs to output human-like responses. \citet{argyle_busby_fulda_gubler_rytting_wingate_2023} showed that by properly conditioning LLMs with targeted identity profiles, they produce biased outputs that strongly correlate with human responses. 
Furthermore, \citet{simmons-2023-moral} claimed that LLMs are moral mimics: by giving models a political identity, they produce texts mirroring the associated moral biases. 
Nevertheless, \citet{santurkar2023whose,hwang2023aligning} discovered that LLMs align poorly with human opinions, as evidenced by model performance using explicit and implicit personae on public opinion polls. We argue that this strategy is suboptimal (\Cref{sec:intro}) due to noisy personae and the inefficient reasoning strategy employed. {\model} overcomes these limitations.

\paragraph{Reasoning with LMs via prompting.} Large-scale model architectures \cite{devlin-etal-2019-bert,radford2019language,gpt3,chowdhery2022palm,touvron2023llama} have enabled LLMs to excel at various reasoning NLP tasks via prompting \cite{wei2022chain,khot2022decomposed,zhou2022least,wang2023selfconsistency,shinn2023reflexion}. Notably, \citet{wei2022chain, kojima2022large} proposed the popular Chain-of-Thought (CoT) techniques, enabling LLMs to explicate intermediate reasoning steps, aiding the solving of multi-step reasoning tasks with higher fidelity and efficiency.

Can CoT analyze and predict human opinion effectively? Surprisingly, a naive application of CoT fails to improve GPT-X models (\Cref{sec:main-results}). We attribute this to the reasoning inconsistencies of CoT and the complexity of the task: strategic reasoning is essential to consistently fully utilize the nuanced explicit and implicit personae. Our VBN reasoning (\Cref{sec:method}) overcomes CoT's limitations. 

\section{LLMs are Distracted by Irrelevant Personae} \label{sec:llms-distracted}

\begin{table}
\footnotesize
\centering
\scalebox{.73}{
\begin{tabular}{l|c|c|ccc}
\toprule
& \textbf{Type} & \textbf{Relevant} & \textbf{+1 less/irr} & \textbf{+3 less/irr} & \textbf{+all less/irr}  \\ 
&  & \textbf{persona} & \textbf{relevant} & \textbf{relevant} & \textbf{relevant} \\ \midrule
\multirow{3}*{\rotatebox{0}{\textbf{ChatGPT}}} & \emph{Explicit} & 37.56 & 35.53\textcolor[RGB]{250,0,0}{$\downarrow$} & 34.51\textcolor[RGB]{250,0,0}{$\downarrow$} & 34.35\textcolor[RGB]{250,0,0}{$\downarrow$} \\ 
& \emph{Implicit} & 34.35 & 33.84\textcolor[RGB]{250,0,0}{$\downarrow$} & 30.76\textcolor[RGB]{250,0,0}{$\downarrow$} & 31.28\textcolor[RGB]{250,0,0}{$\downarrow$} \\ 
\midrule
\multirow{3}*{\rotatebox{0}{\textbf{LLaMa 3.1}}} & \emph{Explicit} & 26.39 & 24.36\textcolor[RGB]{250,0,0}{$\downarrow$} & 23.85\textcolor[RGB]{250,0,0}{$\downarrow$} & 22.84\textcolor[RGB]{250,0,0}{$\downarrow$}  \\ 
& \emph{Implicit} & 26.66 & 25.12\textcolor[RGB]{250,0,0}{$\downarrow$} & 21.02\textcolor[RGB]{250,0,0}{$\downarrow$} & 23.08\textcolor[RGB]{250,0,0}{$\downarrow$}  \\ 
\bottomrule
\end{tabular}}
\caption{\small{The performances of ChatGPT (\emph{gpt-3.5-turbo-0125}) and LLaMa3 (\emph{8B 3.1-it}) significant drop (approx. 1-6\%) on Gun topic of OpinionQA \cite{santurkar2023whose} when less or irrrelevant personae added.}}
\label{fiq:motivation-results}
\end{table}

We study the sensitivity of LLMs to irrelevant personae which motivates \model{} in \Cref{sec:method}. 
This exploration is related to \cite{shi2023large} but we quantify the LLM sensitivity to personae instead. 
We find that LLMs can be easily distracted by explicit or implicit personae that are less or irrelevant to opinion questions. To examine this, we perform a semi-human evaluation to assess the relevancy of personae on 197 randomly chosen Guns questions from the OpinionQA dataset \cite{santurkar2023whose}. Each sample is denoted as $\{T, E, I, q, o, a\}$ where $T$, $E$, and $I$ indicate the topic, explicit personae (demographics and ideology), and implicit personae (historical opinions) of the user answering $q$ with opinion options $o$ and correct label $a$. 

To assess the relevance of each explicit persona in $E$ to $q$, we employ two native English undergraduates and ChatGPT (\emph{gpt-3.5-turbo-0125}) \cite{openai2022chatgpt}. The annotators carefully examined ${q, a}$ and labeled 12 attributes of $E$ as relevant or irrelevant, determined by majority vote. The annotators achieve a good agreement of 60.2\% Krippendorff's alpha \cite{krippendorff2011computing}. For implicit personae ($n = \sim 20$), assessing their relevance to $q$ by manual means is costly. Therefore, we compute the semantic similarity between them and $q$ using OpenAI \emph{text-embedding-ada-002} and label the top-8 as relevant and the rest as irrelevant. 

We test four different setups for each type of personae: (i) using only relevant ones; (ii) including one, (iii) three, or (iv) all irrelevant ones. To ensure that our results are consistent, we use Self-Consistency \cite{wang2023selfconsistency} with $5$ times sampling. We experiment with both representative closed and open source LLMs: ChatGPT and LLaMa3 \cite{touvron2023llama}.

\paragraph{Results in \Cref{fiq:motivation-results}.} Surprisingly, adding a single irrelevant explicit persona results in a prediction change of $30\%$ for ChatGPT and $40\%$ for LLaMa, with $2\%$ performance drop for both. Meanwhile, adding one irrelevant implicit causes much smaller drops for both. Explaining this phenomenon, we have two hypotheses: first, both models rely more on explicit personae for opinion prediction; second, the so-called irrelevant implicit persona may still hold some relevance, as low semantic similarity does not entirely equate to low relevance (Appx. \Cref{fiq:order-disagreement-example}). Additionally, we observe that adding three/all irrelevant for both persona types significantly reduces performance by over 3-5\% in absolute, indicating that irrelevant personae harm model predictions significantly. 

\section{\model: A Chain of Opinion Framework} \label{sec:method}

These above findings suggest that for pre-trained LLMs, the choice of (relevant) personae as input is important and significantly impacts the model's outcomes. Notably, we observe that \emph{training} language models sees even greater benefits when incorporating only relevant personae. We introduce a simple and cost-efficient framework, termed {Chain-of-Opinion (\model{})}  characterizing explicit and implicit personae for opinion prediction. \model{} serves as an intermediate data preprocessing step for fine-tuning and prompting. We assume \emph{optional} access to the user's explicit and implicit personae. Let ${G}_\mathcal{M}: \mathcal{V}^* \to \mathcal{V}^*$ be the generation function of $\mathcal{M}$ where $\mathcal{V}$ is the model vocabulary, $Q$ be the concatenation of the test question and its answer choices, and $a$ be the correct label.

\subsection{\model{} in Prompting} \label{ssec:coo-prompting}

\begin{figure}
\centering
\includegraphics[width=.45\textwidth]{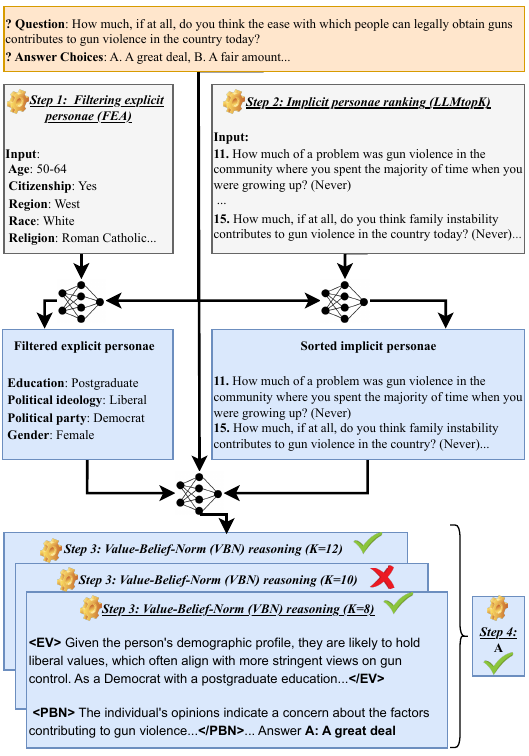}
  \caption{
  {\model} overview with four main steps marked with the nuts. It processes explicit and implicit personae parallelly to facilitate the missing personae scenarios.}
\label{fig:model}

\end{figure}

As introduced in \Cref{sec:intro} and depicted in \Cref{fig:model}, \model{} distinctly processes explicit and implicit personae, motivated by the Value--Belief--Norm theory \cite{stern1999value}, and consists of four main steps: Step 1. Filtering explicit personae; Step 2. Ranking implicit personae; Step 3. Value-Belief-Norm (VBN) reasoning; and Step 4. Answer consistency with dynamic numbers of opinions. 


\paragraph{Step 1. Filtering explicit personae (FEA).}


This step aims to binarily filter explicit personae (the set $E$, denoted in \Cref{sec:llms-distracted}), including user demographics and ideology, for prediction. Irrelevant ones can harm the model performance significantly (\Cref{sec:llms-distracted}), possibly because of LLM's attention mechanism forcing the model to attend to all input tokens, including irrelevant ones. Since the relevancy of an explicit persona to a test question is apparent, we binarily filter out irrelevant explicit personae by instructing the LLM $\mathcal{M}$ to reason and analyze how each of them is helpful for the model to predict the opinion via Chain-of-Thought \cite{wei2022chain}: 

\begin{equation}\label{eq:eq1}
    E_{rel} := G_\mathcal{M}([E, Q])
\end{equation}

The instruction for $\mathcal{M}$ is skipped in \Cref{eq:eq1}. Surprisingly, LLMs evaluate more than half of the explicit personae as irrelevant on average. We conduct human evaluations to examine this in \cref{sec:experimentation}, and provide an example in full in \Cref{appdx:fea-example} showing when considering all explicit personae, the model yields an incorrect prediction while removing unnecessary personae, it offers a correct one.

\paragraph{Step 2. Implicit personae opinions ranking (LLMtop-$K$).}
This step focuses on ranking and selecting implicit personae ($I$) consisting of user historical opinions for prediction. Identifying the most supportive ones for a test question is critical yet complex and often requires significant comprehension efforts. \citet{hwang2023aligning} utilized semantic-similarity between ($q$, $i$) to select implicit persona $\forall i \in I$. This strategy is suboptimal because the top semantic similarity opinions may not be the ones that provide the most supportive information for the models (\Cref{appdx:example-disagreement}, also we hypothesized in \Cref{sec:llms-distracted}). As LLMs are shown to be good data analysts \cite{wang-etal-2023-chatgpt,cheng-etal-2023-gpt}, we propose to utilize LLMs to analyze and rank the implicit personae opinions in usefulness order:

\begin{equation}\label{eq:eq2}
    I_{rel} := top_K(G_\mathcal{M}([I, Q]))
\end{equation}

The instruction of $\mathcal{M}$` is skipped in \Cref{eq:eq2}. We fix $K$ when selecting top-$K$ implicit personae rather than having LLMs directly select useful ones like \model{}'s Step 1 because implicit personae involve more complex and nuanced information with varying degrees of relevance, unlike the binary relevance of explicit personae (\Cref{sec:intro}). Additionally, we input $I$ to LLMs in a \emph{random order} to enhance the versatility of our method. We validate this approach in \Cref{sec:discussion}. Our method prioritizes supportiveness in reasoning about opinions, unlike conventional demonstration selection focusing on semantic similarity or diversity \cite{xu2024context}.


\paragraph{Step 3. Value-Belief-Norm reasoning (VBN).} 
\emph{How can we best utilize $E_{rel}$ and/or $I_{rel}$ for opinion prediction?} Popular prompting methods for reasoning, such as few-shot and zero-shot Chain-of-Thought (CoT) \cite{wei2022chain, kojima2022large}, typically instruct LLMs to reason step-by-step. However, for opinion prediction, a general step-by-step guide presents two key issues. First, the generated reasoning steps can be inconsistent, causing divergent outcomes, particularly at medium-high decoding temperatures ($\geq$ 0.6) (proven in \Cref{sec:method-analysis} and \Cref{appdx:example-cot-inconsistent}). This significantly undermines the \emph{reliability and accuracy} of models. Second, step-by-step reasoning lacks \emph{interpretability}, as it remains unclear how models process and prioritize multiple explicit and implicit personae. When numerous opinions and personae are input, it is difficult to discern which were used in forming the prediction and which were disregarded.


Motivated by the Value-Belief-Norm (VBN) theory \cite{stern1999value}, which outlines the causal relationship chain \{human pro-environmental values $\rightarrow$ beliefs $\rightarrow$ norms $\rightarrow$ behaviors\}, we introduce VBN reasoning: the LLM sequentially analyzing personae in $E_{rel}$ and $I_{rel}$ to derive two variables, \textbf{Environmental Values} from explicit $E_{rel}$ and \textbf{Personal Beliefs and Norms} from implicit $I_{rel}$, and the LLM then predicts the opinion $\hat{a}$ based on these EV and PBN analyses:

\begin{align}\label{eq:eq:3}
    \text{Env. Values} &\rightarrow \text{Per. Beliefs and Norms} \\
    &\rightarrow \hat{a} := G_\mathcal{M}([E_{rel}, I_{rel}, Q])
\end{align}

\Cref{eq:eq:3} is achieved in a chain of thoughts:

\vspace{-2mm}
\begin{tcolorbox}[sharp corners, colback=white, boxrule=-1pt]
\small{$(I_1)$ \texttt{Analyze the user's demographics and ideology one by one to infer their social and environmental values.}}\\
\small{$(I_2)$ \texttt{Analyze the user's historical opinions one by one to infer their beliefs and norms from their social and environmental values.}}\\
\small{$(I_3)$ \texttt{Which opinion is the user likely to choose?}}
\end{tcolorbox}
\vspace{-2mm}

\Cref{fig:model} shows an example of VBN reasoning. Note that the instruction $(I_1)$/$(I_2)$ is skipped if $E_{rel}$/$I_{rel}$ is unavailable. Our human evaluation in \Cref{subsec:human-evaluation-discussion} confirms LLMs are capable of reasoning about human values, beliefs, and norms. This reasoning offers two notable advantages. First, for each question, we ensure that the model explains and analyzes the provided personae one by one thoroughly without omitting any, resulting in more accurate predictions. Second, this method helps the model to output more consistent reasoning explanations, enhancing model reliability (\Cref{sec:method-analysis}).  

\paragraph{Step 4. Answer consistency with dynamic numbers of opinions.}
By fixing $K$ in the $LLMtop-K$ step, we observe that models such as GPT-4 \cite{openai2023gpt4} may refuse to answer the question (\Cref{tab:dynamic-demonstrations}). We attribute this to insufficient implicit personae provided. Inspired by Self-Consistency (SC) \cite{wang2023selfconsistency}, our approach involves sampling multiple answers using different $K$ values for a given question. The most frequent answer, along with the explanation of the first correct answer, becomes the final prediction. Our method is distinct from SC which samples multiple answers with a fixed prompt. We only use three values of K  $\{8,10,12\}$ for efficiency.   

\subsection{\model{} in Fine-tuning} \label{ssec:coo-finetuning}

During fine-tuning, we adopt \model{}'s Steps 1, 2, 3 in \Cref{ssec:coo-prompting} to create \textbf{COO data}. Each sample is denoted as \{$E_{rel}, I_{rel}, \text{Env. Values}, \text{Beliefs and Norms}, Q, a$\} where the model learns to predict $a$ given the rest variables. Since these steps require LLMs with strong instruction-following capabilities, we use ChatGPT as the COO data processor. 

\section{Experiment} \label{sec:experimentation}

\subsection{Experimental Settings}


\paragraph{Dataset.}
We experiment on OpinionQA dataset  \cite{santurkar2023whose} --- the only open-sourced opinion QA dataset to date consisting of both user explicit and implicit personae designed for the assessment of alignment between LMMs' opinions and human participants, encompassing a diverse range of $60$ US demographic groups. We also note the MRFP dataset \cite{sun-etal-2023-measuring}, which, due to privacy concerns, is unavailable for public access.


\paragraph{Dataset preprocessing.} 
Due to limited resources, we follow \citet{hwang2023aligning} to sample a subset of OpinionQA for evaluation. We randomly select $25$ users per topic for our experiments. We use $20\%$ of the implicit questions for each user as the implicit persona. For the remaining $80\%$ implicit questions, we randomly select a maximum of $15$ implicit questions for testing. Our sampling method results in a total of $375$ users and $5,603$ implicit evaluation question--answer pairs. Our subset is highly representative because we gather users from every topic and rigorous statistical tests further validate the significance of our results.

\paragraph{Baselines.}
For prompting experiments, we use both closed- and open-source LLMs: \textbf{ChatGPT} \cite{openai2022chatgpt}, \textbf{ChatGPT-it} \cite{openai2023chatgptinstruct}, \textbf{GPT-4} \cite{openai2023gpt4}, and \textbf{Mistral-7B-it-v.02} \cite{jiang2023mistral} where each model performs all the \model{}'s steps. We compare {\model} with $5$ prompting methods: \textbf{W/o persona}, where LLMs are evaluated without user historical opinions, ideology, or demographic data;  \textbf{Demographic + Ideology + top8 Opinions} (\textbf{DIO-top8}), introduced by \citet{hwang2023aligning} which achieves state-of-the-art results on OpinionQA at that time; \textbf{DIO-top8 + CoT} is the Chain-of-Thought (CoT) prompting \cite{kojima2022large} version of {DIO-top8} by appending \texttt{"answer the following question step-by-step"} to prompts; \textbf{DIO-top8 + SC} is when we apply the Self-Consistency \cite{wang2023selfconsistency} with CoT to \emph{DIO-top8}; \textbf{DIO-top8 + Self-refine} \cite{madaan2023selfrefine} interactively feedbacks and refines the answers by LLMs. 
For GPT-4, we only run the main experiment and use ChatGPT for FEA and LLMtop-$K$ steps due to our limited budget. We provide all the prompt templates and cost analysis in \Cref{appdx:prompts-and-costs}, implementation details in \Cref{appdx:baselines-implementation-details}, and more baselines in \Cref{appdx:more-baselines}.

For fine-tuning, we use ChatGPT to perform \model{}'s steps 1, 2 ($K=8$), 3 as noted in \Cref{ssec:coo-finetuning} on a training set of $30,000$ samples randomly selected from OpinionQA which are different from our $5,603$ test ones. We then fine-tune and evaluate \textbf{GPT-2 models} (base, large) \cite{radford2019language} and \textbf{FlanT5 models} (base, large) \cite{chung2022scaling}. The details are shown in \Cref{appdx:baselines-implementation-details}.

\begin{table*}[t!]
\centering
\scalebox{.635}{
\begin{tabular}{l|ccc|cccc}
\toprule
 &  & \textbf{Prompting} &  & & \textbf{Fine-tuning} & & \\
\textbf{Model} & \textbf{ChatGPT} & \textbf{ChatGPT-it}  & \textbf{Mistral-7B-it-v0.2} & \textbf{GPT-2-base} & \textbf{GPT-2-large} & \textbf{FlanT5-base} & \textbf{FlanT5-large} \\
\midrule
{W/o personae} & 46.60 / 65.72 & 44.91 / 63.60 & 41.24 / 59.54 & \textcolor{red}{\textbf{36.28 / 52.62}} & 21.94 / 39.11 & 48.98 / 68.33 & 39.83 / 58.43 \\
\hdashline
{DIO-top8} & 50.22 / 69.21 & 51.95 / 71.16  & 44.16 / 62.47 & 21.23 / 38.64 & 24.94 / 42.22 & 55.00 / 74.98 & 54.94 / 74.79 \\
{+ Self-refine} & 43.14 / 65.33 & 42.71 / 62.98 & 36.23 / 55.06 & \cellcolor{gray!30} & \cellcolor{gray!30} & \cellcolor{gray!30} & \cellcolor{gray!30} \\
{+ CoT} & 49.96 / 69.05 & 51.90 / 71.51 & 52.25 / 71.95 & \cellcolor{gray!30} & \cellcolor{gray!30} & \cellcolor{gray!30} & \cellcolor{gray!30} \\
{+ CoT-SC} & 50.58 / 69.66 & 52.06 / 71.87 & 53.14 / 72.88 & \cellcolor{gray!30} & \cellcolor{gray!30} & \cellcolor{gray!30} & \cellcolor{gray!30} \\
\hdashline
\emph{+ FEA (Step 1)} & 50.64 / 69.85 & 52.63 / 72.30 & 44.99 / 64.09 & 22.62 / 40.97 & 25.65 / 45.21 & 55.78 / 75.34 & 58.77 / 77.26 \\
\emph{+ VBN (Step 3)} & 51.38 / 70.32 & 52.61 / 71.90 & 53.59 / 73.46 & 24.79 / 43.50 & 28.73 / 47.09 & 58.21 / 77.80 & 56.87 / 76.18 \\
\midrule
\emph{DIO-LLMtop8 (Step 2)} & 51.03 / 70.31 & 52.80 / 72.60 & 45.86 / 64.98 & 22.65 / 41.12 & 28.86 / 47.60 & 57.97 / 77.46 & 58.20 / 77.56 \\
\emph{+ FEA (Step 1)} & 51.19 / 70.69 & 52.97 / 72.84 & 45.23 / 64.73 & 25.05 / 44.41 & 29.54 / 48.66 & 57.45 / 77.13 & 59.00 / 78.46 \\
\emph{+ FEA + VBN} & 52.16 / 71.90 & 53.08 / 72.92 & 54.56 / 74.37 & \textbf{26.17} / \textbf{45.92} & \textbf{30.21} / \textbf{49.63} & \textbf{59.62} / \textbf{78.87} & \textbf{60.13} / \textbf{78.92}\\
\midrule
\textbf{\model{} (ours)} & \textbf{52.66$\dagger$ / 72.75$\dagger$} & \textbf{53.58$\dagger$ / 73.80$\dagger$} & \textbf{54.40$\dagger$ / 74.26$\dagger$} & \textbf{26.17}$\dagger$ / \textbf{45.92}$\dagger$ & \textbf{30.21}$\dagger$ / \textbf{49.63}$\dagger$ & \textbf{59.62}$\dagger$ / \textbf{78.87}$\dagger$ & \textbf{60.13}$\dagger$ / \textbf{78.92}$\dagger$\\
{\% w/ best baseline} & \textcolor{inc}{+ 4.11 / + 4.43} & \textcolor{inc}{+ 2.91 / + 2.68} & \textcolor{inc}{+ 2.37 / + 2.57} & \textcolor{inc}{+ 23.26 / + 18.84} & \textcolor{inc}{+ 21.13 / + 17.55} & \textcolor{inc}{+ 8.40 / + 5.18} & \textcolor{inc}{+ 9.45 / + 5.52} \\
\bottomrule
\end{tabular}
}
\caption{\small{ Accuracy (Acc) / Collapsed Accuracy (CAcc) experimental results. FEA, LLMtop8, and VBN are \model{}'s Steps 1 (explicit), 2 (implicit), and 3 (reasoning). $\dagger$ denotes our method outperforms baselines with p-value < 0.01 under t-test (\Cref{tab:t-test-results}).}} 
\label{tab:overall-results}
\end{table*}

\paragraph{Automatic metrics.} 
We employ \textbf{Accuracy (Acc)} and \textbf{Collapsed Accuracy (CAcc)}\footnote{is a relaxed accuracy wherein the choices of MCQ questions ($\geq$ 4 choices) are collapsed to become 2 choices.} as the evaluation metrics following \citet{hwang2023aligning}. Note that Precision/Recall/F1 is not applicable in our task, since the numbers of answer choices are not the same for all the OpinionQA samples. 

\paragraph{Human metrics.} Human evaluations are crucial due to the absence of automated metrics assessing LLMs' performance in FEA, LLMtop-$K$, and VBN steps of {\model}. Therefore, we conduct our human assessments to address these research questions: \emph{(1) LLMs' effectiveness in filtering unnecessary explicit personae}; \emph{(2) LLMs' proficiency in ranking implicit personae opinions}; \emph{(3) LLMs' ability to explain answers via VBN}. We randomly select $100$ answers generated by {\model} with ChatGPT, ChatGPT-Instruct, GPT-4, and Mistral. We then hire $3$ excellent undergraduates who are native English speakers as annotators. For FEA and LLMtop-$K$ steps, each annotator is instructed to rate on a 1-3 scale (3 is best) via the \textbf{Satisfaction} criterion defined as how well the algorithm of LLMs performs in filtering/ranking, subjectively. To answer (3), we use two criteria named \textbf{Reasonableness} measuring how well the LLMs reason with the VBN explanations, and \textbf{Follow the Instruction} assessing the capability of LLMs in following our instruction to explain and predict the opinions. Three annotators are also guided to rate the criteria on a 1-3 scale. Each metric's final score is the average of three annotators' scores. The scoring instructions are in \Cref{appendix:human-rating-system} and the inter-annotators' agreement is assessed by Krippendorff's alpha \cite{krippendorff2011computing}.

\subsection{Main Results} \label{sec:main-results}

We outline the prompting, fine-tuning, fine-grained, and human evaluation results of \model{}.



\paragraph{Prompting results.}
\Cref{tab:overall-results} shows our main experimental outcomes. For GPT-4, it attains 57.98\% Acc with DIO-top8 and 59.42\% with \model{}, establishing a strong SOTA result surpassing the previous of 53.74\% \cite{hwang2023aligning}. Overall, \model{} delivers the best results with significant improvements of 2-4\% Acc for all benchmarked LLMs with ChatGPT securing the most gain of 4.11\%. Its component steps consistently enhance baselines with VBN  > LLMtop8 > FEA on average. 

Among prompting methods, we observe that na\"{\i}ve CoT helps Mistral slightly but harms ChatGPT and ChatGPT-it, while SC improves all. We attribute this to the inconsistency and unreliability of CoT reasoning (\ref{sec:method}), and the challenge of this task. Meanwhile, \model{}'s VBN reasoning consistently improves, verifying the effectiveness of requiring LLMs to explicitly analyze all the personae through values, beliefs, and norms. Additionally, self-refine consistently lowers model performance, indicating that multiple refinement rounds may be counterproductive for this complex task as these rounds may amplify the model's inherent biases \cite{xu2024perils}, leading to more biased predictions.


Among models, notably, for GPT-4, we use ChatGPT for FEA and LLMtop-$K$ steps, showcasing the strength of a weaker model that enhances a stronger one. Finally, ChatGPT, ChatGPT-it, and Mistral show improvements by selecting only $4.79/12$ and $5.59/12$, $8.83/12$ explicit personae on average. This suggests that over half of explicit personae are noisy for model opinion prediction.

\paragraph{Fine-tuning results.} 
\Cref{tab:overall-results} presents our fine-tuning results. Overall, leveraging \model{}'s FEA, LLMtop-$K$, and VBN extra variables results in significant improvements for both decoder-only and encoder-decoder models, with average gains of 22.20\% for GPT-2s and 8.93\% for FlanT5s. Among \model{}'s steps, FEA contributes the least, while LLMtop-$K$ and VBN extra variables drive more substantial gains across most models.

The VBN extra variables can be seen as distilled knowledge from ChatGPT, intuitively enhancing fine-tuning results. However, notably, \model{}'s FEA and LLMtop-$K$, which focus on selecting relevant explicit and implicit personae, already deliver substantial improvements across all models, bringing FlanT5-large's performance on par with GPT-4. This verifies our hypothesis that removing irrelevant personae is necessary for high performance. 

Finally, GPT-2-base performs surprisingly well even without user demographics and ideology, possibly due to potential contamination \cite{sainz-etal-2023-nlp} with public polling data from OpinionQA.

\paragraph{Fine-grained Results.} We compare \model{} with the DIO-top8 baseline across 15 OpinionQA topics to assess its performance in detail. Overall, we see \model{} consistently outperforms DIO-top8 in most topics, with the largest improvements in ``View on gender'' (+17.69\% with Mistral), ``Autonomous vehicles'' (+13.49 with GPT-2), and ``Misinformation'' (+11.61 with ChatGPT). These gains further emphasize \model{}'s effectiveness in enhancing LM performance on social and belief-driven topics, especially those involving complex reasoning. Full results are provided in Appx. \Cref{tab:fine-grained-results}.

\paragraph{Human evaluation results.} \label{subsec:human-evaluation-discussion}

\begin{table}[t!]
\centering
\scalebox{.65}{
\begin{tabular}{l|c|c|cc}
\toprule
\textbf{Model} & \textbf{FEA Satis.} & \textbf{LLMtopK Satis.} &  \textbf{VBN Rea.} & \textbf{VBN FI}\\
\midrule
ChatGPT & $2.56_{\alpha = 0.74}$ & $\textbf{2.32}_{\alpha=0.68}$ & $2.85_{\alpha = 0.88}$ & $\textbf{2.91}_{\alpha = 0.90}$\\
ChatGPT-it & $\textbf{2.64}_{\alpha=0.71}$ & $2.28_{\alpha=0.65}$ & $2.87_{\alpha=0.90}$ & $\textbf{2.95}_{\alpha=0.87}$ \\
GPT-4 & \cellcolor{gray!30} & \cellcolor{gray!30} & $\textbf{2.90}_{\alpha=0.91}$ & $2.21_{\alpha=0.77}$\\
Mistral & $2.31_{\alpha=0.65}$ & $2.12_{\alpha=0.64}$ & $2.58_{\alpha = 0.68}$ & $2.16_{\alpha=0.55}$ \\
\bottomrule
\end{tabular}
}
\caption{\small{Human evaluation results. LLMs perform adequately in \model{}'s FEA step, and excel in VBN reasoning, but face challenges with the LLMtop-K step. \llong{$\alpha$ denotes the Krippendorff's alpha.}}}
\label{tab:huam-evaluation-results}
\end{table}

For \model{}'s FEA and LLM ranking steps, from \Cref{tab:huam-evaluation-results}, ChatGPT and ChatGPT-it generally achieve similar performance and are better than Mistral: ChatGPT excels slightly in ranking while ChatGPT-it performs slightly better in performing FEA. Three models are proficient in FEA but struggle with the ranking task where the common error is misplacing relevant opinions, due to this task's complexity. Second, four models effectively generate VBN reasoning thoughts, and GPT-4 performs the best.  Finally, ChatGPT and ChatGPT-it follow our instructions to explain and analyze the explicit and implicit personae provided one by one with VBN significantly better than GPT-4 and Mistral, achieving nearly perfect scores of $3$. Our hypothesis is they are optimized for following instructions, while  GPT-4 is optimized for completing texts.


\section{Discussion}\label{sec:discussion}
We discuss the main analyses here, including \model{} generalization (\Cref{ssec:coo-generalization}) and \model{}'s Steps sequentially. Extra analyses are supplemented in \Cref{appdx:extra-analysis}.

\begin{table}[t!]
\centering
\scalebox{.7}{
\begin{tabular}{l|c}
\toprule
\textbf{Method} & \textbf{ChatGPT} \\
\midrule
W/o personae & 46.60 \\
\emph{\model{} w/o personae (Step 4  activated)} & \textbf{47.79} \\
\midrule
DIO-top8 w/o explicit & 49.22  \\
\emph{\model{} w/o explicit (Steps 2, 3 (PBN), 4 activated)} & \textbf{51.66} \\
\midrule
DIO-top8 w/o implicit & 47.16 \\
\emph{\model{} w/o implicit (Steps 1, 3 (EV), 4 activated)} & \textbf{50.13} \\
\bottomrule
\end{tabular}
}
\caption{\small{\model{}'s results with missed persona(e) with ChatGPT. In all scenarios, \model{} outperforms the baselines significantly.}}
\label{tab:generalization-results}
\end{table}

\subsection{\model{} in Missing Personae Scenarios} \label{ssec:coo-generalization}
\model{} demonstrates strong generalizability even when explicit, implicit, or both types of personae are missing. Specifically, in the absence of both, \model{} reduces to Self-consistency \cite{wang2023selfconsistency}; without explicit personae, only Step 1 of \model{} is skipped; without implicit personae, Step 2 is omitted. As shown in \Cref{tab:generalization-results}, \model{} consistently outperforms the leading baseline by an absolute margin of 2-3\% in all scenarios. Finally, its steps can be generalizable to other tasks, see \Cref{appdx:generalization-to-other-tasks}.


\subsection{Method Analysis} \label{sec:method-analysis}

\paragraph{FEA: ablation study.} 
To gauge the impact of removing irrelevant explicit personae (FEA), we experiment with applying FEA exclusively to the baseline DIO-top8 \cite{hwang2023aligning}, denoted as DIO-top8 + FEA in \Cref{tab:overall-results}. We observe a 1-2\% Acc performance boost on ChatGPT, ChatGPT-it, and Mistral respectively. This underscores the effectiveness of eliminating irrelevant explicit personae in improving the model prediction.

\paragraph{FEA: irrelevant personae distribution.} 
To understand the explicit personae filtered by LLMs across various topics, we document the top 3 removed personae in \Cref{appdx:top3-attributes-removed}. {"Citizenship"} is observed to be the most frequently removed attribute, followed by {"Race"}. This could be due to LLMs treating these as sensitive information, prioritizing respect and unbiased text generation. Another explanation may be the lack of correlation between citizenship/race and opinions in the OpinionQA dataset. Additionally, we also see that ChatGPT often categorizes {``Marital status"} as non-useful, ChatGPT-it commonly removes {``Frequency of religious attendance"}, and {``Gender"} got removed by Mistral, revealing potential biases in LLMs.

\paragraph{LLMtop-$K$: compared to semantic top-$K$.} 
From \Cref{tab:overall-results}, DIO-LLMtop8 outperforms DIO-top8 by $1-4\%$ accuracy on ChatGPT, ChatGPT-it, and Mistral, confirming that prioritizing usefulness over semantic similarity improves model prediction. 
To further understand the difference between semantic similarity orders and usefulness orders, we discuss (1) the agreement of LLM orders and semantic similarity orders, and (2) maximum disagreement points between these orders. 

In tackling (1), we calculate Kendall's Tau coefficient \cite{kendall_tau} between the orders generated by ChatGPT, ChatGPT-it, Mistrial, and semantic similarity orders, and the results are presented in Appx. \Cref{fig:ranking-agreements}. Surprisingly, for ChatGPT and ChatGPT-it, we find that the two ranking orders have no agreement with means approximating $0$. For Mistral, we observe a low agreement with a mean of $0.43$ score. These low and no agreements further verify that ranking by usefulness can be very different from ranking by semantic similarity. 

For answering (2), \Cref{appdx:example-disagreement} illustrates one such case in the \texttt{"Guns"} topic. We observe that not all top-8 opinions by semantic similarity scores help predict the opinion. For example, the $16$-th opinion, despite having a relatively high semantic similarity score with the question which might offer some perspective on the prevalence of guns in the user's community during the upbringing, is less directly related to the question. This is similar to the $18$-th opinion which is also less relevant. Meanwhile, several important opinions are deselected by the semantic-similarity-based method, such as the $6,3,4,10$-th ones, which are chosen by the LLM. The $6$-th one is critical, and directly relevant because it assesses the person's attitude toward safety measures related to gun ownership. Finally, by using the LLMtop-$K$ order, the model predicts the opinion accurately, whereas the semantic similarity order leads to an incorrect prediction.

\paragraph{LLMtop-$K$: the order of input opinions.} 
We study the (1) sensitivity and the (2) performance variance of LLMs to the order of input implicit personae in the LLMtop-$K$ step. 

To address (1), our discovery confirms sensitivity, but with reasonable overlap when $K$ is sufficiently large ($K\geq8$). We randomly select $300$ questions, shuffle implicit persona opinions four times with different seeds, and record four LLM ranking outputs for each. We also collect one more LLM ranking output by feeding implicit personae opinions in semantic similarity order. For each $K\in\{1,2,...,20\}$, we calculate the pairwise Overlap Coefficient (OC) \cite{vijaymeena2016survey} among the five ranking outputs, averaging them as the {LLM ranking consistency score} for each $K$. The scores, shown in Appx. \Cref{fiq:llmtop-consistency-score}, indicate that for $K\geq8$, the ranking outputs overlap well with a score of $\geq .6$ for both models. 

For (2), we find no significant performance variance. Specifically, we assess ChatGPT and Mistral with DIO-LLMtop8 on $3$ out of $4$ random seeds, detailed in \Cref{appdx:ranking-consistency}. The results demonstrate relatively small standard deviations in their performance, and critical values of $99\%$ CI of DIO-LLMtop8 under t-test for both models surpass DIO-top8, confirming that LLMtop8's effectiveness is not due to randomness.

\paragraph{VBN: compared to CoT.} 

\Cref{tab:overall-results} indicates that Chain-of-Thought (CoT) \cite{kojima2022large} slightly harms the performance for ChatGPT and ChatGPT-it. Conversely, our Value-Belief-Norm (VBN) reasoning enhances performance for all models. To investigate the consistency of CoT and VBN, we design an experiment with ChatGPT, DIO-top8 where we randomly select $100$ question-answer pairs and sample $5$ answers per pair using CoT and VBN, at $3$ different temperatures $0.3$, $0.6$, $0.9$. We measure the percentage of questions that all $5$ answers sampled have the same result, as the consistency score. The results are illustrated in Appx. \Cref{fig:consistency} showing that VBN brings better consistent answers compared to CoT, especially when the temperature is high verifying VBN potentially enhances the reliability of LLMs.   

\paragraph{Answer consistency with dynamic opinions.} 
We study (1) how frequently LLMs are unable to answer the question and (2) the impact on performance when more than $K=8$ opinions are provided. \Cref{tab:dynamic-demonstrations} provides the results. We find that with $8$ opinions, GPT-4 exhibits the highest percentage of unanswered questions, while Mistral answers all the questions. Increasing the \#opinions beyond $8$ reduces this percentage across models, confirming our hypothesis regarding the lack of implicit personae opinions when fixing $K=8$ in \Cref{sec:method}. Lastly, while increasing $K$ could harm the model performance, \model{}'s answer consistency enables LLMs to achieve the best results across K values. 

\begin{table}[t!]
\centering
\scalebox{.55}{
\begin{tabular}{l|c|c|c|c}
\toprule
\textbf{Model} & \textbf{ChatGPT} & \textbf{ChatGPT-it} & \textbf{GPT-4} & \textbf{Mistral}\\
\midrule
\emph{ITA of DIO-LLMtop8 + FEA + VBN} & 0.20 & 0.91 & 3.40 & 0.00 \\
\emph{DIO-LLMtop8 + FEA + VBN} & 52.16 &  53.08 & 59.11 & 54.56 \\
\midrule
\emph{ITA of DIO-LLMtop10 + FEA + VBN} & 0.00 & 0.00 & 1.43 & 0.00 \\
\emph{DIO-LLMtop10 + FEA + VBN} & 51.89 & 52.90 & 58.88 & 53.62 \\
\midrule
\emph{ITA of DIO-LLMtop12 + FEA + VBN} & 0.00 & 0.00 & 0.00 & 0.00 \\
\emph{DIO-LLMtop12 + FEA + VBN} & 51.60 & 52.03 & 59.18 & 54.21 \\
\midrule
\textbf{\model} & \textbf{52.66} & \textbf{53.58} & \textbf{59.42} & \textbf{54.40} \\
\bottomrule
\end{tabular}
}
\caption{\small {Percentage of ``Impossible To Answer'' (ITA) (\%) observed with corresponding performance during generation.}}
\label{tab:dynamic-demonstrations}
\end{table}

\section{Conclusion} 
This paper identifies two major challenges in aligning LLMs with human opinions via personae: noisy personae and ineffective reasoning strategies over personae. To address these, we propose {\model}, a novel four-step framework in the light of the Value-Belief-Norm theory: the first two steps tackle the noise issue using LLMs as analysts, while the last two enhance reasoning through the novel VBN reasoning. {\model} significantly improves LLM prompting and fine-tuning, demonstrating high generalizability in scenarios with missing personae and potential applications in other personalization tasks.

\section*{Acknowledgements}
\llong{This research project is partially supported by the National Research Foundation Singapore under the AI Singapore Programme (AISG Award No: AISG2-TC-2023-010-SGIL), the Singapore Ministry of Education Academic Research Fund Tier 1 (Award No: T1 251RES2207), and the National Research Foundation, Singapore under its AI Singapore Programme (AISG Award No: AISG2-GC-2022-005). Do Xuan Long is supported by the A*STAR Computing and Information Science (ACIS) Scholarship.} 

\section*{Limitations} One limitation of {\model} is that it requires the LLMs to be well capable of following human instructions to solve tasks such as selecting explicit personae, ranking historical opinions, and explaining personae and opinions VBN reasoning. However, we foresee that this limitation is going to be overcome by cutting-edge AI language models, in the present and near future. Additionally, \model{} utilizes user's personal information from explicit and implicit personae, which may be sensitive to some audiences and not be fully available in the real world. However, to what extent is the personal information provided, our {\model} is still able to offer reasonable opinion predictions since it is not constrained by the number of provided explicit personae, or the number of user historical opinions (see \Cref{ssec:coo-generalization}). \llong{Finally, for fine-tuning, \model{} currently leverages ChatGPT for data synthesis, which may pose challenges to replicability. To address this, we will open-source our generated data and code \href{https://github.com/dxlong2000/COO}{here}. Future research could explore using open-source LLMs, which are increasingly powerful and comparable to proprietary models.} 

\section*{Ethical Considerations}
Characterizing and predicting human opinions with LLMs can be directly applied to personalize and align machines to users’ values, and cultural beliefs. Nonetheless, there exist unwanted situations when LLMs with our techniques can be misused for unethical purposes and biased opinions. 

\paragraph{Bias amplification and fairness.} A personalized LLM allows users to reinforce their existing beliefs and potentially amplify biased or unethical perspectives, leading to the creation of echo chambers \cite{del2016echo}. This can ultimately harm users by reinforcing polarized or undesirable views. To mitigate this issue, the Chain-of-Opinion (CoO) reasoning from our proposed {\model} involves presenting user demography or ideology group responses alongside personalized answers. Additionally, \model{} can encourage users to reflect on their previous viewpoints.

\paragraph{Privacy and consent.} Users may not always be aware of or have control over the extent of personalization applied to the content they receive. Therefore, empowering users to have control over AI-generated opinions is essential. Users should be able to customize and adjust the explicit and implicit personae used for opinion prediction. This customization can help mitigate potential biases and provide individuals with AI-generated opinions that align more closely with their values and preferences.

\paragraph{Using ChatGPT for data synthesis.} \llong{We follow prior studies \citep{RAY2023121,tan-etal-2024-large} to use OpenAI ChatGPT to synthesize data. Additionally, we obey OpenAI's terms of use\footnote{https://openai.com/policies/row-terms-of-use/} to use ChatGPT's synthesized data to develop models that do not compete with OpenAI.} 

\paragraph{Misuse and responsibility for long-term societal impact.} \llong{While our method aims to align opinions with individuals, it also introduces risks of misuse, such as the propagation of harmful ideologies or manipulating human opinions. While this is not what our method is designed for, there is no way to prevent this type of misuse. We emphasize the importance of ethical alignment in deploying these systems and suggest that developers and practitioners must establish robust guidelines and oversight mechanisms to prevent misuse. Furthermore, incorporating mechanisms to monitor and audit AI-generated content following ethical norms is crucial.}

\llong{We also recognize potential risks of unintended societal consequences, such as fostering group biases or undermining collective decision-making processes. We also acknowledge the potential for unintended societal consequences, such as fostering group biases or undermining collective decision-making processes. To mitigate these risks, we recommend integrating ethical safeguards, including mechanisms to monitor and audit AI use, and fostering user awareness and diverse perspectives during model training and deployment. This approach aims to minimize negative societal impacts and ensure the technology is applied constructively and equitably.}

\paragraph{Human evaluation.} Through human evaluations, we observe that our proposed method does not generate any discriminatory, insulting responses. We validate the intermediate steps of our proposed {\model} by human evaluation which involves manual labor. We hire annotators to score, and the hourly pay is set to $\$20$, which is higher than the local statutory minimum wage. Therefore, we do not anticipate any major ethical concerns raising from human evaluations.



\bibliography{anthology,custom}

\newpage

\appendix
\section{Implementation Details} \label{appdx:baselines-implementation-details} 

\paragraph{Prompting.} ChatGPT \emph{(gpt-3.5-turbo)}, ChatGPT-it \emph{(gpt-3.5-turbo-instruct)}, GPT-4 \emph{(gpt-4)} are called via OpenAI API with chat, text, text completion mode respectively at a temperature of $0.3$. Mistral-7B-Instruct-v0.2 is called via HuggingFace interface\footnote{\url{https://huggingface.co/mistralai/Mistral-7B-Instruct-v0.2}}. We use Nucleus Sampling \cite{Holtzman2020The} with a $p=.95$ as our decoding strategy. To obtain the embeddings of opinions for semantic similarity scores' computations, we use OpenAI's \emph{text-embedding-ada-002} model with its default setting, following \citet{hwang2023aligning}. For each sample, {\model} requires $5$ inference calls, $2$ for FEA and LLMtop-$K$ steps, and $3$ for $K\in\{8,10,12\}$. Therefore, to have a fair comparison with our method, we sample $5$ answers for the Self-Consistency baseline, and $2$ rounds of feedback-edit for Self-refine baseline, for each question.

\paragraph{Fine-tuning.} We fine-tune GPT-2 \cite{radford2019language} and FlanT5 \cite{chung2022scaling} base and large models to verify that \model{}'s Steps 1, 2, 3 (\Cref{sec:method}) also help to build better opinion-aligned models. Both models with two different sizes are initialized from public pre-trained checkpoints on the Transformers library \cite{wolf-etal-2020-transformers} of HuggingFace. We use a learning rate of $1e-5$ for FlanT5, and $5e-5$ for GPT-2, and AdamW \cite{loshchilov2018decoupled} as our optimizer with a warm-up of 100 steps. FlanT5 variants are trained on $50K$ iterations, and evaluations and checkpoint-savings are done for each $1000$ steps. GPT-2 base model is trained on $15$ epochs and evaluated every 300 steps, while GPT-2 large is trained on only $5$ epochs, and the checkpoints are evaluated every 300 steps. All the models are fine-tuned on a single A100 80GB GPU. We use a window size of $1024$ for both models, and Nucleus Sampling \cite{Holtzman2020The} with a $p=.95$ as our decoding strategy, same as API/inference models. The input format for both models is \texttt{``Input: explicit\_persona <SEP> implicit\_persona <SEP> EV <SEP> PBN <SEP> question <SEP> answer\_choices; Output: correct\_answer"} for with persona cases, and \texttt{``Input: question <SEP> answer\_choices; Output: correct\_answer"} for without persona case. The \texttt{``correct\_answer"} is an actual text correct answer like \texttt{``Yes/No"}, unlike API/inference models where we use \texttt{``A/B/C/D"}. We find that fine-tuning with the textual correct answer yields significantly better results compared to \texttt{``A/B/C/D"}, while prompting with \texttt{``A/B/C/D"} for API/inference models achieve slightly better results compared to textual output.

\section{Additional Analyses} \label{appdx:extra-analysis}

\subsection{Fine-grained Analyses} \label{subsec:fine-grained-analysis}

\begin{table*}[t!]
\begin{subtable}{1\textwidth}
\centering
\scalebox{.63}{
\begin{tabular}{l|ccccc|c}
\midrule
& \textbf{Guns} & \textbf{Auto. vehicles} & \textbf{Views on gender} & \textbf{Community types} & \textbf{Race} \\
\midrule
\textbf{ChatGPT} & 53.87 / \textbf{57.06} & 45.33 / \textbf{51.25} & 53.21 / \textbf{59.23} & \textbf{43.47} / 40.88 & 43.06 / \textbf{43.27}\\ 
\textbf{ChatGPT-it} & 57.00 / \textbf{58.21} & 44.78 / \textbf{51.92} & 52.15 / \textbf{53.07} & 45.24 / \textbf{46.14} & 44.65 / \textbf{48.28} \\
\textbf{Mistral} & 44.73 / \textbf{58.12} & 41.72 / \textbf{53.75} & 40.09 / \textbf{57.78} & 35.45 / \textbf{42.08} & 41.11 / \textbf{51.44} \\
\textbf{GPT-4} & 60.39 / \textbf{63.37} & \textbf{53.22} / 50.00 & 63.73 / \textbf{71.43} & 42.86 / \textbf{47.96} & \textbf{55.17} / 50.57 \\
\hdashline
\textbf{GPT-2} & \textbf{27.96} / 27.73 & 16.83 / \textbf{30.32} & 16.40 / \textbf{27.35} & 14.33 / \textbf{27.00} & 22.14 / \textbf{29.59} \\ 
\textbf{GPT-2 large} & 25.34 / \textbf{30.07} & 26.04 / \textbf{31.96} & 22.66 / \textbf{24.66} & 23.33 / \textbf{29.00} & 21.67 / \textbf{22.42} \\
\textbf{FlanT5} & \textbf{62.08} / 60.39 & 55.60 / \textbf{59.61} & 57.45 / \textbf{62.60} & 45.98 / \textbf{47.20} & 54.23 / \textbf{61.56} \\
\textbf{FlanT5 large} & 60.56 / \textbf{65.45} & 51.52 / \textbf{58.87} & 60.28 / \textbf{60.64} & 46.21 / \textbf{49.76} & 49.24 / \textbf{55.82} \\
\midrule
& \textbf{Gender \& Leadership} & \textbf{America in 2050} & \textbf{Trust in science} & \textbf{Bio. \& food issues} & \textbf{Misinformation} \\
\midrule
\textbf{ChatGPT} & 48.27 / \textbf{52.22} & 46.93 / \textbf{49.46} & 54.93 / \textbf{56.43} & 52.27 / \textbf{54.75} & 49.33 / \textbf{60.94} \\ 
\textbf{ChatGPT-it} & 54.70 / \textbf{56.28} & 46.20 / \textbf{49.00} & \textbf{61.58} / 55.50 & 55.86 / \textbf{57.26} & 52.11 / \textbf{53.62} \\
\textbf{Mistral} & 50.23 / \textbf{57.87} & 35.14 / \textbf{47.60} & 51.65 / \textbf{60.37} & 52.78 / \textbf{58.58} & 50.77 / \textbf{53.85} \\
\textbf{GPT-4} & \textbf{65.55} / 63.03 & \textbf{53.71} / 45.27 & 61.54 / \textbf{68.46} & 58.03 / \textbf{61.61} & 52.71 / \textbf{57.26} \\
\hdashline
\textbf{GPT-2} & \textbf{29.82} / 23.50 & 25.73 / \textbf{29.27} & 21.01 / \textbf{30.96} & 19.03 / \textbf{27.93} & 14.33 / \textbf{26.71} \\ 
\textbf{GPT-2 large} & 21.50 / \textbf{28.11} & 27.90 / \textbf{29.27} & 29.61 / \textbf{30.96} & 28.60 / \textbf{33.52} & 23.34 / \textbf{30.13} \\
\textbf{FlanT5} & 63.98 / \textbf{66.08} & \textbf{55.82} / 55.00 & \textbf{64.22} / 63.00 & 61.41 / \textbf{61.56} & \textbf{60.49} / 60.00 \\
\textbf{FlanT5 large} & 58.54 / \textbf{64.00} & 46.54 / \textbf{51.30} & 61.76 / \textbf{68.43} & 57.34 / \textbf{63.75} & 51.77 / \textbf{61.97} \\
\midrule
& \textbf{Privacy \& Surveilance} & \textbf{Family \& Relationships} & \textbf{Economic inequality} & \textbf{Global attitudes} & \textbf{Political views} & \textbf{Average} \\
\midrule
\textbf{ChatGPT} & 53.24 / \textbf{54.29} & 57.22 / \textbf{61.00} & 45.60 / \textbf{52.43} & \textbf{49.60} / 45.54 & \textbf{56.97} / 51.15 & 50.22 / \textbf{52.66} \\ 
\textbf{ChatGPT-it} & 51.02 / \textbf{54.33} & 57.89 / \textbf{58.25} & \textbf{51.98} / 50.13 & 57.23 / \textbf{57.86} & 46.85 / \textbf{53.84} & 51.95 / \textbf{53.58} \\
\textbf{Mistral} & 43.31 / \textbf{58.06} & 47.42 / \textbf{58.50} & 41.87 / \textbf{51.89} & 42.0 / \textbf{52.76} & 44.13 / \textbf{53.34} & 44.16 / \textbf{54.40} \\
\textbf{GPT-4} & 47.73 / \textbf{52.27} & 62.50 / \textbf{63.89} & 63.81 / \textbf{64.76} & \textbf{66.67} / 63.58 & 62.07 / \textbf{67.82} & 57.98 / \textbf{59.42} \\
\hdashline
\textbf{GPT-2} & 17.45 / \textbf{35.75} & \textbf{32.44} / 29.17 & 18.29 / \textbf{28.21} & 21.49 / \textbf{25.29} & 21.29 / \textbf{23.07} & 21.23 / \textbf{26.17} \\ 
\textbf{GPT-2 large} & 29.53 / \textbf{37.30} & 24.19 / \textbf{28.57} & 22.28 / \textbf{28.21} & 24.87 / \textbf{32.01} & 23.38 / \textbf{28.46} & 24.94 / \textbf{30.21} \\
\textbf{FlanT5} & \textbf{58.36} / 58.00 & 64.59 / \textbf{65.60} & 54.06 / \textbf{57.60} & \textbf{63.23} / 58.04 & 49.08 / \textbf{52.80} & 55.00 / \textbf{59.62} \\
\textbf{FlanT5 large} & 57.92 / \textbf{61.53} & 64.98 / \textbf{67.02} & 51.66 / \textbf{58.72} & 57.19 / \textbf{57.95} & 48.55 / \textbf{56.67} & 54.94 / \textbf{60.13} \\
\bottomrule
\end{tabular}
}
\end{subtable}
\caption{\small{Fine-grained accuracy results of models with DIO-top8 \cite{hwang2023aligning} / Chain-of-Opinion (\model{}; ours).}}
\label{tab:fine-grained-results}
\vspace{1mm}
\end{table*}

\Cref{tab:fine-grained-results} presents our fine-grained results across topics of the OpinionQA dataset \cite{santurkar2023whose}. Overall, \model{} consistently outperforms DIO-top8 across most of the OpinionQA topics. The highest absolute improvements occur in ``View on gender'' (+17.69 with Mistral), ``Autonomous vehicles'' (+13.49 with GPT-2), and  ``Misinformation'' (+11.61 with ChatGPT). {These improvements highlight the \model{}'s strength in helping models better handle complex, belief-based topics, particularly those involving social and political biases.} 

Across models, \model{} notably enhances performance, whether in strong models like GPT-4 or weaker ones like GPT-2. GPT-4, one of the leading LLMs for reasoning, shows marked improvement in scientific and social topics such as ``Views on gender'' (+7.70) and ``Misinformation'' (+4.55). GPT-2, usually less effective, also benefits from \model{}'s enhancements, with significant gains in ``Views on gender'' (+11.37) and ``Political views'' (+7.98). This further emphasizes the strength of \model{} in enhancing both high-performing and less capable models.

\subsection{Additional Baselines} \label{appdx:more-baselines}

\begin{table}[t!]
\centering
\scalebox{.6}{
\begin{tabular}{l|cc}
\toprule
\textbf{Model}  & ChatGPT & Mistral-7B-Instruct-v0.2 \\
\midrule
DIO-top8 & 50.22 & 44.16 \\
\midrule
DIO-top8 + FEA & 50.64 & 44.99 \\
DIO-top8 + Random FEA (S=2000) & 49.47 & 42.23 \\
DIO-top8 + Random FEA (S=2024) & 48.85 & 43.36 \\
\midrule
DIO-LLMtop8 & 51.03 & 45.86 \\
DIO + Random LLMtop8 (S=2000) & 48.13 & 44.58 \\
DIO + Random LLMtop8 (S=2024) & 49.21 & 43.84 \\
\bottomrule
\end{tabular}
}
\caption{\small{Accuracy results of ChatGPT and Mistral with two trivial variants with two different random seeds 2000 and 2024 in \Cref{appdx:more-baselines}.}}
\label{tab:additional-random-baselines}
\end{table}

We compare \model{}'s FEA and LLMtop-$K$ steps with two simple variants outlined in \Cref{tab:additional-random-baselines}. Given ChatGPT and Mistral's strong performance with just $4.79/12$ and $8.83/12$ explicit persona attributes, a crucial question arises: \emph{(1) Can comparable performance be achieved by randomly selecting $5/12$ and $9/12$ explicit persona attributes instead of relying on LLMs?}. Our answer is no. The first variant, \emph{DIO-top8 + Random FEA}, involves \emph{randomly selecting} $5/12$ and $9/12$ explicit persona attributes. The second variant entails \emph{randomly selecting} $8$ implicit persona opinions instead of using \model{}'s LLMtop-$K$ step. From \Cref{tab:additional-random-baselines}, we find that randomly selecting explicit persona attributes significantly harms the performance of both models due to the removal of important attributes. Additionally, randomly selecting $8$ implicit persona opinions also adversely affects the models, particularly ChatGPT. These observations underscore the effectiveness and importance of \model{}'s FEA and LLMtop-$K$ steps.

\subsection{Top-3 Removed Explicit Personae Attributes} \label{appdx:top3-attributes-removed}
\Cref{tab:removed-attributes-statistics} reveals a relatively common pattern in how three LLMs, ChatGPT, ChatGPT-it, and Mistral-7B-Instruct-v0.2, filter explicit personae across various topics. \texttt{"Citizenship"} and \texttt{"Race"} consistently emerge as the most frequently removed attributes, suggesting a deliberate effort by these models to minimize potential biases associated with these demographic factors. Additionally, ChatGPT-it’s tendency to remove \texttt{"Frequency"} and Mistral-7B-Instruct's broader removal of \texttt{"Education"} and \texttt{"Political Party"} highlight model-specific strategies and comprehension in filtering personae based on topic relevance. Overall, these patterns suggest that not all explicit personae are equally relevant for predicting opinions.

\begin{table*}[t!]
\centering
\scalebox{.43}{
\begin{tabular}{l|l|l|l}
\toprule
\textbf{Topic} & \textbf{ChatGPT} & \textbf{ChatGPT-Instruct} & \textbf{Mistral-7B-Instruct-v0.2}\\
\midrule
Guns & {'Citizenship', 'Race', 'Marital status'} & {'Citizenship', 'Frequency of religious attendance', 'Religion'} & {'Citizenship', 'Education', 'Religion'} \\
Automation \& driverless vehicles & {'Citizenship', 'Race', 'Marital status'} & {'Citizenship', 'Race', 'Frequency of religious attendance'} & {'Citizenship', 'Religion', 'Frequency of religious attendance'} \\
Views on gender & {'Citizenship', 'Race', 'Frequency of religious attendance'} & {'Citizenship', 'Race', 'Frequency of religious attendance'} & 'Citizenship', 'Religion', 'Frequency of religious attendance'  \\
Community types \& sexual harassment & {'Citizenship', 'Race', 'Gender'} & {'Citizenship', 'Frequency of religious attendance', 'Race'} &  'Education', 'Race', 'Political Party'\\
Biomedical \& food issues & {'Citizenship', 'Race', 'Marital status} & {'Citizenship', 'Race', 'Marital status'} & 'Citizenship', 'Race', 'Marital status' \\
Gender \& Leadership & {'Citizenship', 'Race', 'Region'} & {'Citizenship', 'Race', 'Frequency of religious attendance'} & 'Region', 'Race', 'Citizenship' \\
America in 2050 & {'Citizenship', 'Race', 'Marital status'} & {'Citizenship', 'Race', 'Frequency of religious attendance'} & 'Citizenship', 'Frequency of religious attendance', 'Race'\\
Trust in science & {'Citizenship', 'Marital status', 'Race'} & {'Citizenship', 'Race', 'Marital status'} & ’Citizenship’, ’Race’, 'Region' \\
Race & {'Citizenship', 'Marital status', 'Age'} & {'Citizenship', 'Age', 'Religion'} & 'Marital status', 'Education', 'Age'\\
Misinformation & {'Citizenship', 'Marital status', 'Race'} & {'Citizenship', 'Marital status', 'Race'} & 'Citizenship', 'Race', 'Religion' \\
Privacy \& Surveillance & {'Citizenship', 'Race', 'Marital status'} & {'Citizenship', 'Race', 'Frequency of religious attendance'} & 'Religion', 'Race', 'Region'\\
Family \& Relationships & {'Citizenship', 'Race', 'Region'} & {'Citizenship', 'Race', 'Frequency of religious attendance'} & 'Citizenship', 'Race', 'Religion'\\
Economic inequality & {'Citizenship', 'Frequency of religious attendance', 'Race'} & {'Citizenship', 'Frequency of religious attendance', 'Race'} & 'Gender', 'Citizenship', 'Religion' \\
Global attitudes & {'Marital status', 'Race', 'Citizenship'} & {'Citizenship', 'Marital status', 'Race'} & 'Gender', 'Frequency of religious attendance', 'Marital status' \\
Political views & {'Citizenship', 'Marital status', 'Frequency of religious attendance'} & {'Citizenship', 'Frequency of religious attendance', 'Race'} & 'Frequency of religious attendance', 'Gender', 'Citizenship' \\
\bottomrule
\end{tabular}
}
\caption{\small {Top-3 explicit personae that got removed the most by the LLMs.}}
\label{tab:removed-attributes-statistics}
\end{table*}

\begin{figure}[t!]
\centering
\includegraphics[width=.45\textwidth,trim={0cm 0cm 0cm 0cm},clip]{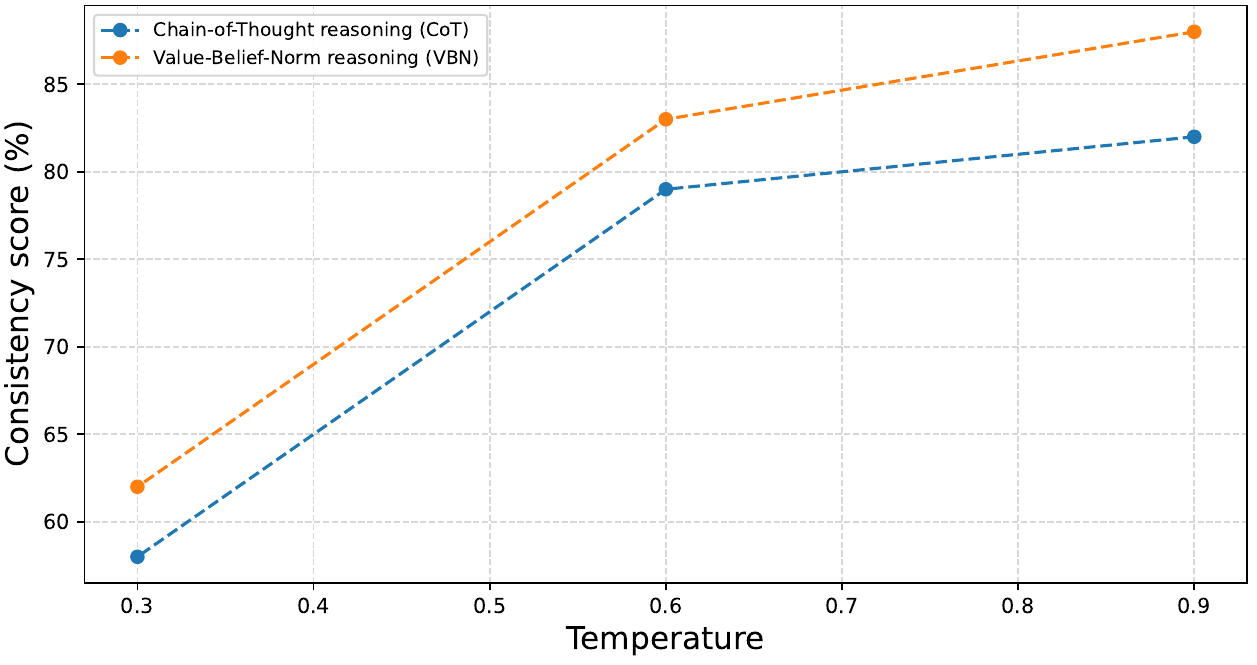}
\caption{\small{Consistency scores of the baseline DIO-top8 (ChatGPT) with CoO and CoT.}}
\label{fig:consistency}
\end{figure} 

\subsection{Consistency and Reliability of VBN versus CoT}\label{appdx:consistency-scores}
\Cref{fig:consistency} presents the consistency scores of the baseline DIO-top8 (ChatGPT) with VBN and CoT reasoning over 100 samples. Both methods show improved consistency as temperature increases, with scores rising from about 58\%-82\% for CoT, and  62\%-88\% for VBN. VBN consistently outperforms CoT across temperatures, suggesting that it is more robust and reliable compared to CoT.







\subsection{Ranking Consistency for LLMtop-$K$ Step} \label{appdx:ranking-consistency}

\begin{figure}
\centering
\includegraphics[width=.47\textwidth,trim={0cm 0cm 0cm 0cm},clip]{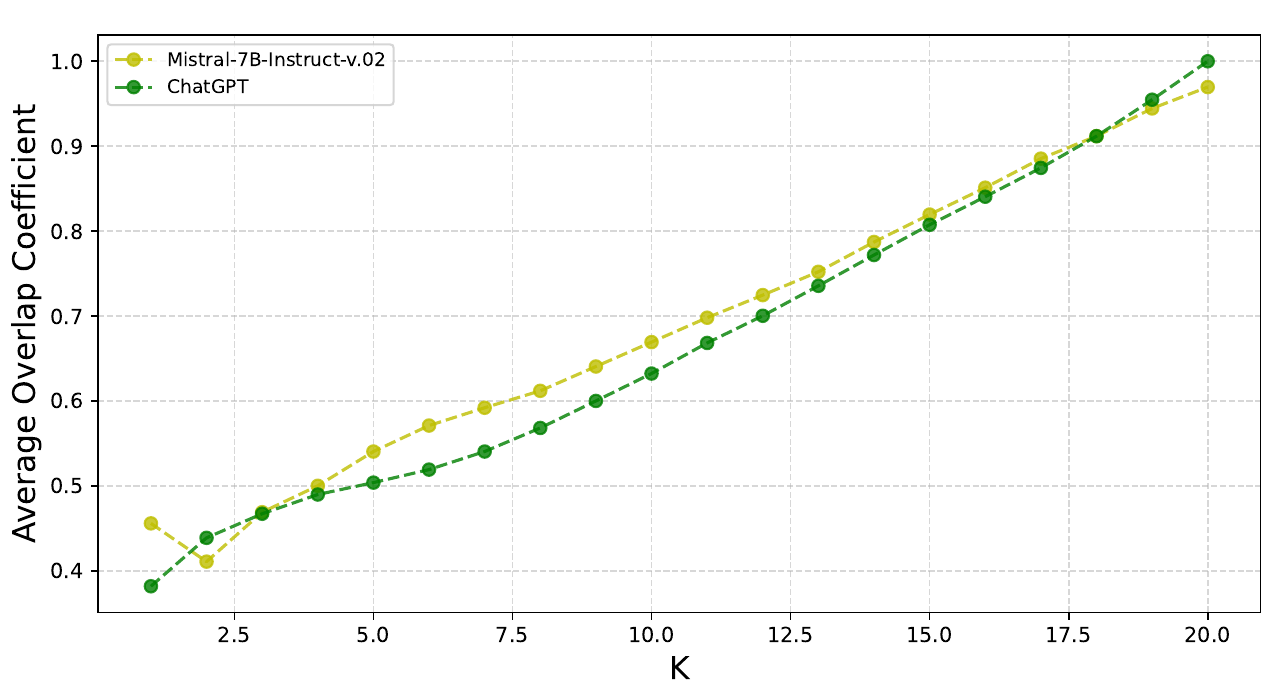}
 \caption{\small{ChatGPT and Mistral-7B-Instruct-v.02 overlap coefficient values for different values of $K$. We observe that for $K$ is large enough ($K \geq 8$), the coefficient value is relatively acceptable ($\geq 0.6$).}}
\label{fiq:llmtop-consistency-score}
\end{figure}

We compute the average pairwise Overlap coefficient \cite{vijaymeena2016survey} $OC(A, B) = \frac{|A \cap B|}{\min(|A|, |B|)}$ across five ranking outputs generated by five input strategies \Cref{fiq:llmtop-consistency-score}. The performance of these strategies, evaluated on 300 random samples, is detailed in \Cref{tab:ranking-consistency-scores}. Results show a negligible variance across three different random seeds, indicating that randomizing the order of implicit personae for the $LLMtop8$ step yields a relatively stable strategy.

\begin{table}[t!]
\centering
\scalebox{.6}{
\begin{tabular}{l|c|ccc|c}
\toprule
\textbf{Model} & \textbf{Method} & \textbf{Seed = 2024} & \textbf{Seed = 5} & \textbf{Seed = 2000} &  \textbf{Std}\\
\midrule
ChatGPT & DIO-LLMtop8  & 51.03 & 50.95 & 51.11  & 0.0652 \\
Mistral & DIO-LLMtop8  & 45.86 & 45.55 & 45.36 & 0.2060 \\
\bottomrule
\end{tabular}
}
\caption{\small {Accuracy results of ChatGPT and Mistral on our test set with DIO-LLMtop8 where different orders of input implicit persona opinions are tested for LLMtop-K step.}}
\label{tab:ranking-consistency-scores}
\end{table}

\subsection{Kendall's Tau Scores for Ranking Agreements}\label{sec:kendall-tau-score}

\begin{figure*}
\centering
\begin{minipage}[b]{0.3\textwidth}
\includegraphics[scale=.23]{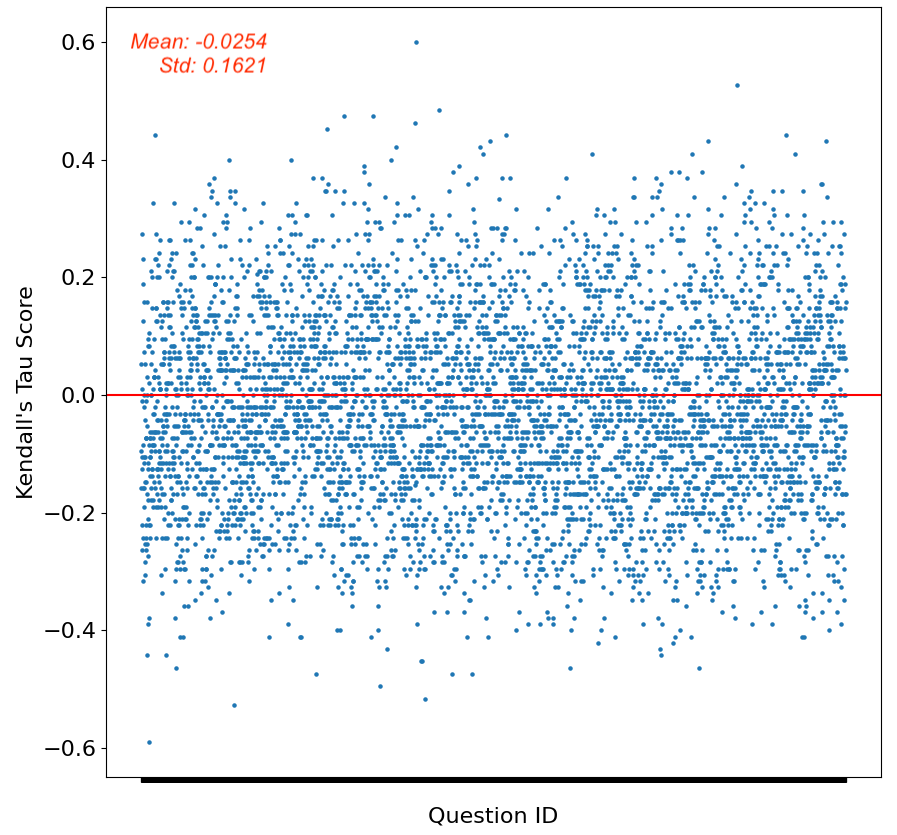}
\end{minipage}
\hfill
\begin{minipage}[b]{0.3\textwidth}
\includegraphics[scale=.23]{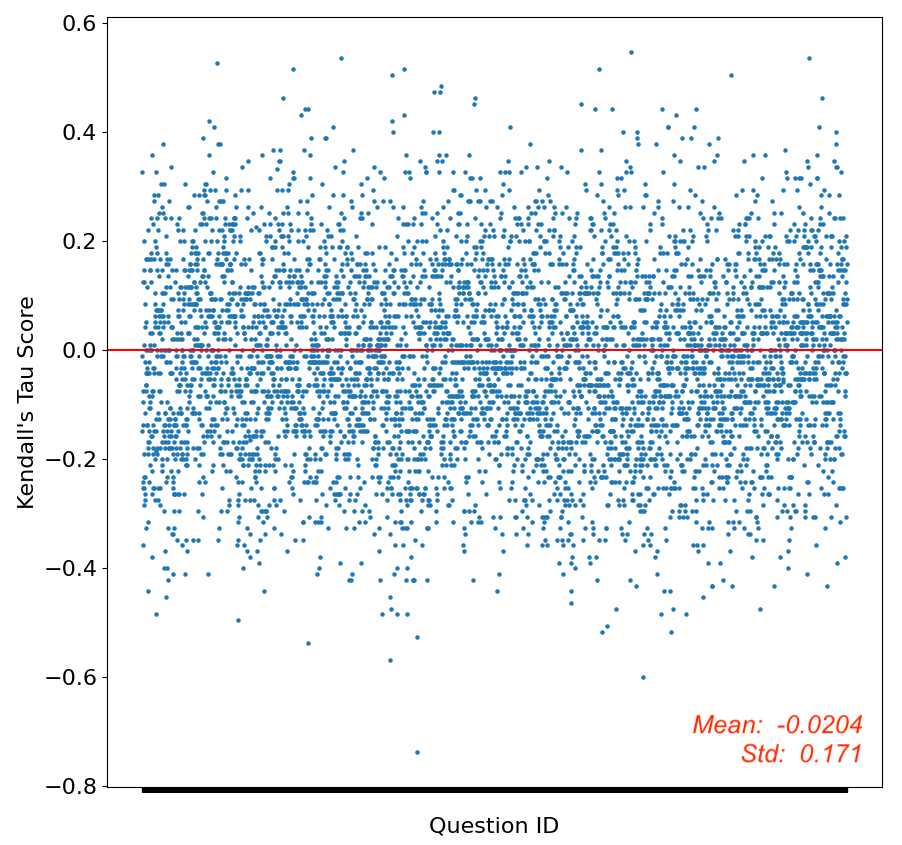}
\end{minipage}
\hfill
\begin{minipage}[b]{0.3\textwidth}
\includegraphics[scale=.2]{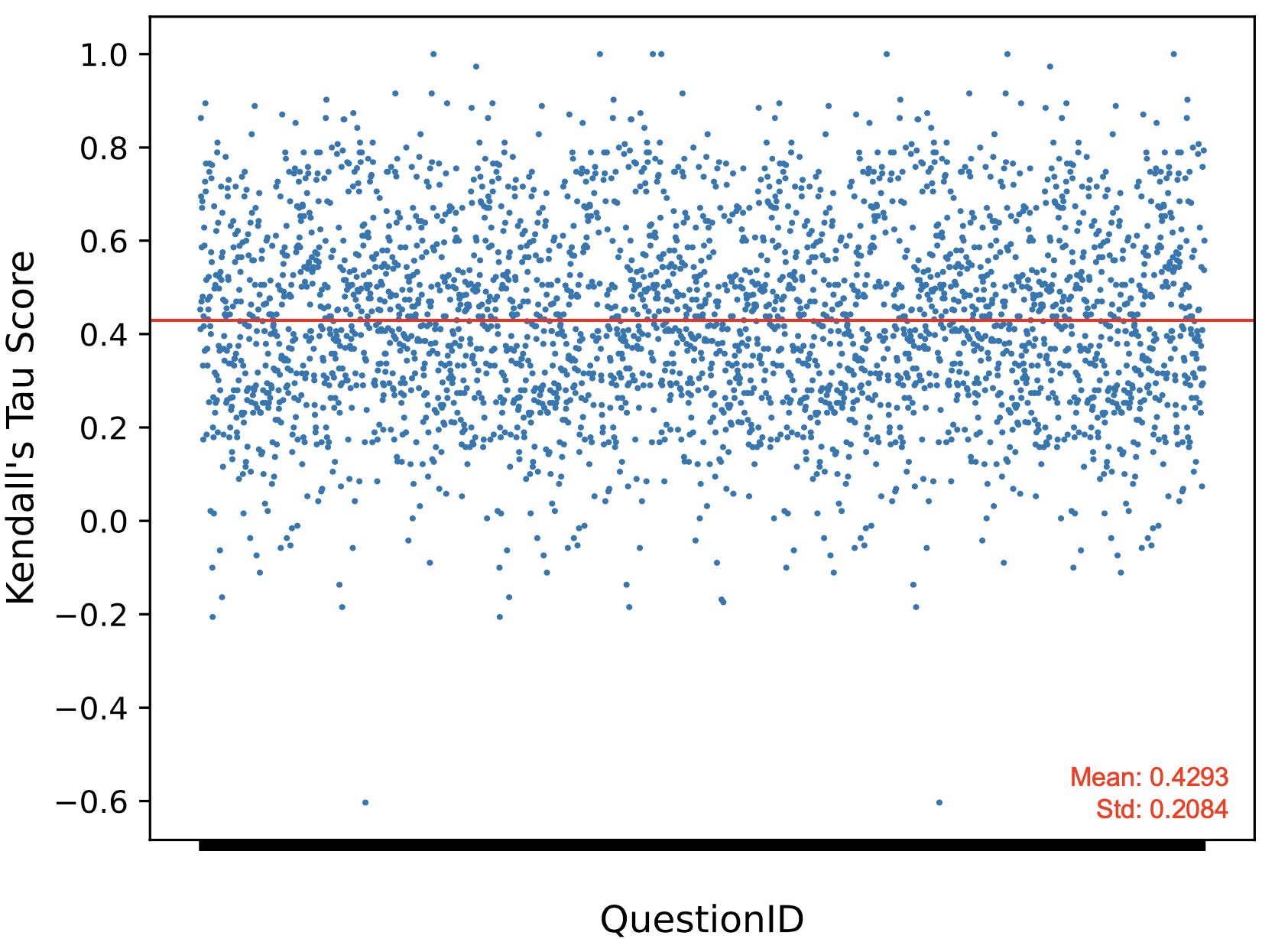}
\end{minipage}
\caption{\emph{Left / Middle / Right:} Ranking agreements between ChatGPT top-$K$ / ChatGPT-it / Mistral orders and semantic similarity orders. One example that has a high disagreement score is shown in \Cref{appdx:example-disagreement}.}\label{fig:ranking-agreements}
\end{figure*}


\Cref{fig:ranking-agreements} shows our ranking agreement between ChatGPT, ChatGPT-it, Mistral orders and semantic similarity orders. We observe that ChatGPT and ChatGPT-it orders have minimal monotonous relations with means approximating 0 and low standard deviations with semantic orders. More specifically, with ChatGPT, the maximum agreement is 0.6000 while the minimum is -0.5895 and the Kurtosis is -0.2173. For ChatGPT-it, the maximum is slightly lower with 0.5473, while the minimum is -0.7368 which is smaller ChatGPT, and the Kurtosis is -0.1017. Meanwhile, Mistral shows a low correlation of 0.43. These low and no correlations highlight that usefulness orders can significantly differ from the semantic similarity orders commonly used in previous studies.




\subsection{Student T-test Results for \Cref{tab:overall-results}}

\begin{table}[t!]
\centering
\scalebox{.8}{
\begin{tabular}{l|c|c}
\toprule
\textbf{Model} & \textbf{Accuracy} & \textbf{Collapsed Accuracy} \\
\midrule
ChatGPT & 4.11e-11 & 6.06e-13 \\
ChatGPT-Inst. & 9.97e-8 & 4.45e-5 \\
GPT-4 & 4.23e-6 & 1.17e-9 \\
Mistral & 6.01e-8 & 4.12e-6 \\
\midrule
GPT-2-base & 2.19e-69 & 1.82e-43 \\
GPT-2-large & 5.62e-73 & 6.09e-49 \\
FlanT5-base & 1.23e-19 & 3.19e-12 \\
FlanT5-large & 2.55e-21 & 1.20e-17
 \\
\bottomrule
\end{tabular}
}
\caption{\small{The p-value computed by student t-test. We observe that all the values are significantly smaller than 0.01 verifying the significance of our improvements.}}
\label{tab:t-test-results}
\end{table}

We employ the Student t-test to assess the statistical significance between \model{} and the best-performing baseline for each model in \Cref{tab:overall-results}. Essentially, under the null hypothesis:
\begin{itemize}
    \item \emph{H0: There is no significant difference}.
    \item \emph{H1: There is a significant difference}.
\end{itemize}

As we can see, the p-values from the tests in \Cref{tab:t-test-results} are significantly below 0.01, indicating the significance of \model{} improvements. 

\subsection{\model{}'s Generalization to Other Tasks} \label{appdx:generalization-to-other-tasks}

Each step of \model{} aligns closely with methodologies from prior studies that have proven effective in personalized tasks. We outline them below:

\begin{itemize}
\item \textbf{Step 1: FEA.} Filtering irrelevant user attributes to improve generation outcomes is well-established \cite{xu2024prompting}. For instance, \citet{rao2011hierarchical,sakaki-etal-2014-twitter} filter gender information using classifiers, while \citet{kim-etal-2017-demographic} focus on age, and \citet{demszky-etal-2019-analyzing} analyze political polarity. Our FEA step is thus generalizable and applicable to these tasks.

\item \textbf{Step 2: LLMtop-K.} Selecting the top-K most relevant historical opinions for the next prediction is conceptually related to re-ranking recommendations using LLMs \cite{hou2024large,xu2024prompting} and selecting key utterances in dialogue generation \cite{do-etal-2022-cohs}. Our LLMtop-K step can likewise enhance personal chat and recommendation tasks.

\item \textbf{Step 3: VBN reasoning.} The Value-Belif-Norm reasoning is an adaptation of Chain-of-Thought prompting \cite{wei2022chain,kojima2022large}. This strategy is designed for opinion prediction but can be used for tasks where user values, beliefs, and norms are crucial features.

\item \textbf{Step 4: Majority voting with a dynamic number of historical opinions.} This step applies to any task where leveraging dynamic demonstrations enhances prediction accuracy.
\end{itemize}

In conclusion, \model{}'s steps are well-supported by prior studies and can be generalized to benefit personalized tasks.

\section{Prompts} \label{appdx:prompts-and-costs}

\subsection{Cost Analyses for API Models}\label{appendix:prompting-cost}
Prompting costs for API models are detailed in \Cref{tab:prompting-cost-analysis}. For GPT-4, \model{} is priced similarly to the baseline DIO-top8, while DIO-top8 + SC costs nearly twice as much. This is because we execute the FEA and LLMtop-$K$ steps of \model{} using ChatGPT, which is comparatively inexpensive. For ChatGPT and ChatGPT-it, \model{} incurs an additional $7$ to $10$ US dollars compared to DIO-top8 + SC. However, this extra cost is justified by the significant performance gains, with particularly large improvements in certain topics.

\begin{table}
\centering
\resizebox{.475\textwidth}{!}{
\begin{tabular}{ccccc|c}
\midrule
 & DIO-top8 & DIO-top8 + CoT & DIO-top8 + SC & \model{} & Model \\
 \midrule
Avg. \#tokens & 562.72 & 623.62  & 995.89 & 3227.18  & ChatGPT \\
Total US\$  & 3.01 & 3.73 & 6.82 & 14.05 & ChatGPT \\
\midrule
Avg. \#tokens & 562.72 & 630.58  & 1019.31 & 3418.72  & ChatGPT-it \\
Total US\$  & 3.12 & 3.84 & 7.11 & 20.11 & ChatGPT-it \\
\midrule
Avg. \#tokens & 559.27 & - & 1021.14* & 3292.66  & GPT-4 \\
Total US\$  & 91.19 & - & 226.15* & 125.60 & GPT-4  \\
\bottomrule
\end{tabular}
}
\caption{\small{Prompting cost analysis of \model{} and other baselines as of 1st Sep 2024. * denotes our estimation on $50$ samples.}}
\label{tab:prompting-cost-analysis}
\end{table}

\subsection{Prompt Template for \model{}'s FEA}
\label{sec:fea-prompts}

\begin{tcolorbox}[fonttitle=\small, fontupper=\small]
\texttt{A person can be described by the following attributes:}

\texttt{\{original\_attribute\_list\}}

\texttt{Based on the above list of demographic information above, now I give you a new question with possible answer choices:}

\texttt{Question: '\{test\_question\}'}

\texttt{Answer choices: '\{test\_choices\}'}

\texttt{Please analyze which attributes in the demographic information are useful for you to answer the above question step by step. Give me the output in the Python list format: [...]}

\texttt{Give me the answer in the format below:}

\texttt{Explanations: ... }

\texttt{Answer: [...]}
\end{tcolorbox}

\subsection{Prompt Template for  \model{}'s LLMtop-K}
\label{sec:reranking-prompts}

\begin{tcolorbox}[fonttitle=\small, fontupper=\small]
\texttt{Given social behavior question-answer pairs answered by a user about his/her opinions about \{subtopic\}:}

\texttt{\{original\_persona\_question\_order\}}

\texttt{You are an expert in analyzing the social behaviors of a user. Given a new question asking him/her:} 

\texttt{'\{test\_question\}'} 

\texttt{Your task is to sort the list of given question-answer pairs in descending order such that the first question-answer pair brings the most useful information to answer the new question, whilst the last question-answer pair brings the least useful information.} 

\texttt{Give me the answer in the form of a Python list of indexes:}

\texttt{Answer: [...]}
\end{tcolorbox}

\subsection{Prompt Template for \model{}'s VBN Reasoning}

\begin{tcolorbox}[fonttitle=\small, fontupper=\small]

\texttt{A person can be described as follows:}

\texttt{\{explicit\_persona\_str\}}

\texttt{The person has the following opinions on \{topic\}.}

\texttt{Opinions:}

\texttt{\{implicit\_persona\_str\}}

\texttt{Given the following question:}

\texttt{Question: \{question\}}

\texttt{Answer choices: \{choice}\}

\texttt{Answer the above question by following the steps below:}

\texttt{Analyze the user's demographics and ideology one by one to infer their social and environmental values. Wrap this analysis by <EV> and </EV>.}

\texttt{Analyze the user's historical opinions to infer their beliefs and norms from their social and environmental values. Wrap this analysis by <PBN> and </PBN>.}

\texttt{From the above analyses, which opinion he is likely to choose? Answer: A. or B. or C. or D. or E....}

\end{tcolorbox}

\subsection{Prompt Templates for Baselines}
\label{sec:prompt-template-baselines}
We use the same prompt templates for ChatGPT \cite{openai2022chatgpt}, ChatGPT-it \cite{openai2023chatgptinstruct}, GPT-4 \cite{openai2023gpt4}. The template prompts for baselines are presented below.

\subsubsection{W/o Persona \cite{santurkar2023whose}}

\begin{tcolorbox}[fonttitle=\small, fontupper=\small]
\texttt{Question: \{question\}}

\texttt{Answer choices:}

\texttt{\{choice\}}

\texttt{Complete the answer by the following format without any explanation:}

\texttt{Answer: A. or B. or C. or D. or E...}
\end{tcolorbox}

\subsubsection{DIO-top8 \cite{hwang2023aligning}}

\begin{tcolorbox}[fonttitle=\small, fontupper=\small]
\texttt{A person can be described as follows:}

\texttt{\{explicit\_persona\_str\}}

\texttt{The person has the following opinions on \{topic\}.}

\texttt{Opinions:}

\texttt{\{implicit\_persona\_str\}}

\texttt{Based on the above information, which answer choice is the user most likely to choose?}

\texttt{Question: \{question\}}

\texttt{Answer choices: \{choice}\}

\texttt{Give the answer in the format:}

\texttt{Answer: A. or B. or C. or D. or E....}
\end{tcolorbox}

\subsubsection{Self-refine \cite{madaan2023selfrefine}}

\begin{tcolorbox}[fonttitle=\small, fontupper=\small]
\texttt{You are given a question and an answer for that question. Analyze the question and the answer and provide some feedback on the answer to the question. Don't change the answer, just provide feedback.}

\texttt{Question: \{test\_question\}}

\texttt{Choices: \{choices\}}

\texttt{Answer: \{selected\_choice\}}

\texttt{Feedback:}
\end{tcolorbox}

\begin{tcolorbox}[fonttitle=\small, fontupper=\small]
\texttt{You are given a question, an answer to that question, and feedback to the answer. Based on the feedback, refine your answer and generate the final answer in around 170 words.}

\texttt{Question: \{test\_question\}}

\texttt{Answer: \{selected\_choice\}}

\texttt{Feedback: \{feedback\}}

\texttt{Refined answer: {new\_choice + explanation}}
\end{tcolorbox}

\subsubsection{Chain-of-Thought \cite{kojima2022large}}

\begin{tcolorbox}[fonttitle=\small, fontupper=\small]
\texttt{A person can be described as follows:}

\texttt{\{explicit\_persona\_str\}}

\texttt{The person has the following opinions on \{topic\}.}

\texttt{Opinions:}

\texttt{\{implicit\_persona\_str\}}

\texttt{Based on the above information, answer the following question step-by-step:}

\texttt{Question: \{question\}}

\texttt{Answer choices: \{choice}\}

\texttt{Give the answer in the format:}

\texttt{Answer: A. or B. or C. or D. or E....}

\texttt{Explanations:...}
\end{tcolorbox}

\section{Human Evaluation}

\subsection{Human Grading Instructions}\label{appendix:human-rating-system}

Our details of human rating instructions are provided in \Cref{tab:human-rating-instruction} for all the criteria. It is worth noting that selecting all features can't get a high FEA Satisfaction score, according to our instructions. In addition, if the selected explicit personae fall among several scores, the annotators are instructed to take the minimum score.

\begin{table*}[t!]
\centering
\scalebox{.6}{
\begin{tabular}{l|l}
\toprule
\textbf{Criterion} & \textbf{Scoring Instruction} \\
\midrule
& 1: The number of filtered-out explicit personae that are directly relevant for answering the question is more than 3. \\
& 1: The number of selected explicit personae that are somewhat irrelevant for answering the question is more than 3. \\
& 2: The number of filtered-out explicit personae that are directly relevant for answering the question is 2 or 3. \\
\textbf{FEA Satisfaction} & 2: The number of selected explicit personae that are somewhat irrelevant for answering the question is 2 or 3. \\
& 3: The number of filtered-out explicit personae that are directly relevant for answering the question is less than or equal to 1. \\
& 3: The number of selected explicit personae that are somewhat irrelevant for answering the question is less than 2. \\

 \midrule
 & 1: Among the top-8 implicit persona opinions, the number of less relevant opinions for answering the question is more than 4.\\
\textbf{LLMtop-$K$ Satisfaction} & 2: Among the top-8 implicit persona opinions, the number of less relevant opinions for answering the question from 2 to 4.\\
 & 3: Among the top-8 implicit persona opinions, the number of less relevant opinions for answering the question is less than or equal to 1.\\
 \midrule
  & 1: The VBN has limited or flawed values, beliefs, norms thoughts with inadequate support. \\
\textbf{VBN Reasonableness} & 2: The VBN has some values, beliefs, norms thoughts with decent support but room for improvement.\\
 & 3: The VBN has strong, clear, and well-supported values, beliefs, norms thoughts with a comprehensive understanding.\\
 \midrule
  & 1: The generated VBN explanation does not mention more than $4$ attributes/opinions from explicit and implicit personae.\\
\textbf{VBN FI} & 2: The generated VBN explanation somewhat follows the instruction by involving more than $4$ attributes/opinions but room for improvement.\\
 & 3: The generated VBN explanation follows perfectly the instructions via explaining all the explicit and implicit attributes one by one.\\
\bottomrule
\end{tabular}
}
\caption{\small {Human rating instructions. FEA, LLMtop-$K$, and CoO stand for Filtering Explicit Personae Attributes, Implicit Personae Opinions Ranking, and Chain-of-Opinion reasoning (\Cref{sec:method}).}}
\label{tab:human-rating-instruction}
\end{table*}

\section{Error Analyses and Examples}

\subsection{Error Analyses} \label{subsec:error-analysis}
\paragraph{FEA misses key explicit personae.} 
Despite showing promising results with FEA, LLMs sometimes misselect relevant personae. One such example is the top-left of \Cref{appdx:error-analysis}. We observe that in this case, our annotators can't grade a high FEA satisfaction score because \texttt{"Education"} and \texttt{"Age"} are also two important personae as they can influence one's understanding of workplace dynamics significantly, which are deselected by ChatGPT.

\paragraph{LLMtop-$K$ opinions consist of less relevant ones.} We observe LLMs frequently include less relevant, or even irrelevant opinions to the ranked list such as in \Cref{appdx:error-analysis}-bottom. We attribute this to the challenge of this task, even for humans it might require substantial cognitive effort.

\paragraph{LLMs may not follow the instructions to perform VBN reasoning.} Although ChatGPT and ChatGPT-it demonstrate a robust ability to perform VBN reasoning (\Cref{tab:huam-evaluation-results}), the same level of proficiency is not observed in Mistral and GPT-4, as exampled in \Cref{appdx:error-analysis}-top-right. We posit this disparity arises from the fact that ChatGPT and ChatGPT-it excel in comprehending and executing human instructions, while GPT-4 excels primarily in generating coherent text.

\subsection{Examples}
\begin{figure}[t!]
\centering
\includegraphics[width=.45\textwidth,trim={0cm 0cm 0cm 0cm},clip]{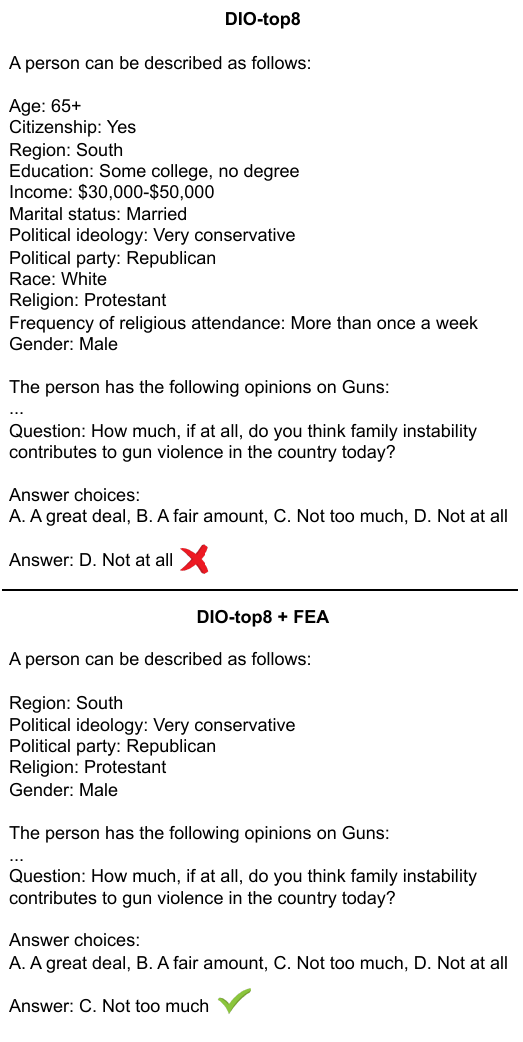}
 \caption{\small{FEA example with ChatGPT.}}
\label{fiq:fea-example}
\end{figure}

\paragraph{FEA example with ChatGPT.}\label{appdx:fea-example}
\Cref{fiq:fea-example} shows an FEA example with ChatGPT. We observe that by removing unnecessary explicit personae including \texttt{"Age"}, \texttt{"Citizenship"}, \texttt{"Education"}, \texttt{"Income"}, \texttt{"Marital Status"}, \texttt{"Race"}, \texttt{"Frequency of religious attendance"}, ChatGPT predicts the opinion accurately, while without removing, an incorrect prediction was made.

\begin{figure*}[t!]
\centering
\includegraphics[width=1\textwidth,trim={0cm 0cm 0cm 0cm},clip]{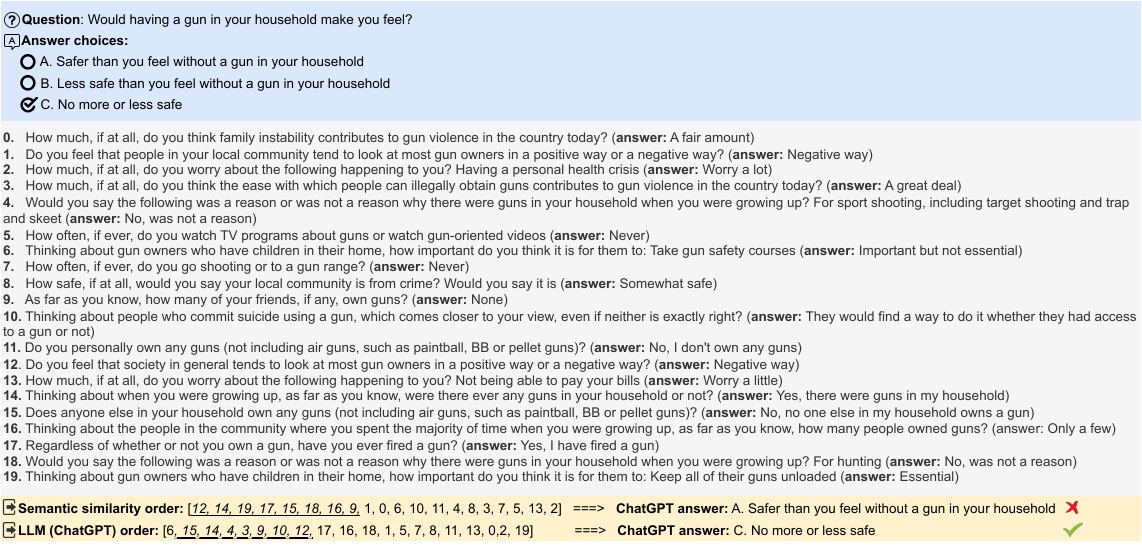}
 \caption{\small{Example of the high disagreement between orders by semantic similarity scores and LLM (ChatGPT).}}
\label{fiq:order-disagreement-example}
\end{figure*}

\paragraph{Example of high disagreement between rankings.} \label{appdx:example-disagreement}
\Cref{fiq:order-disagreement-example} illustrates one example of the high disagreement between orders by semantic similarity scores and LLM (ChatGPT).  We derive three observations, as discussed in \Cref{sec:method-analysis}. First, not all top-8 opinions by semantic similarity scores help predict the opinion. For example, $16$-th opinion, despite having a relatively high semantic similarity score with the question which might offer some perspective on the prevalence of guns in the user's community during the upbringing, is less directly related to the question. This is similar to the $18$-th opinion which is also less relevant. Meanwhile, several important opinions are deselected by the semantic-similarity-based method, such as the $6,3,4,10$-th ones, which are chosen by the LLM. The $6$-th one is critical, and directly relevant because it assesses the person's attitude toward safety measures related to gun ownership. Finally, by using LLMtop-$K$ order, the model predicts the opinion accurately, while an incorrect prediction is made with the semantic similarity order.

\paragraph{Example of inconsistent answers generated by CoT.} \label{appdx:example-cot-inconsistent}

\begin{figure*}[t!]
\centering
\includegraphics[width=.9\textwidth,trim={0cm 0cm 0cm 0cm},clip]{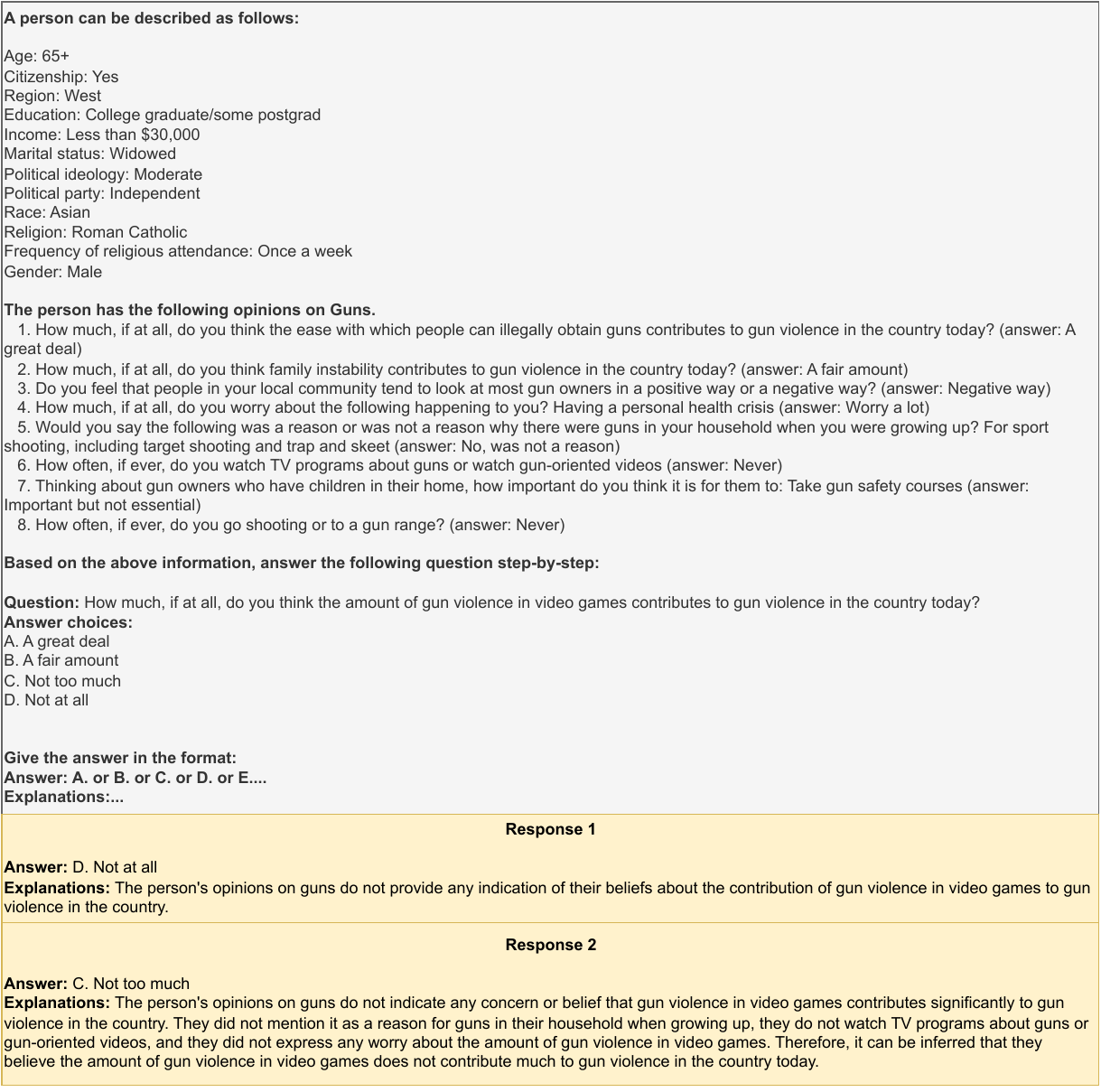}
 \caption{\small{Example of the inconsistent answers generated by ChatGPT with Chain-of-Thought.}}
\label{fiq:example-cot-inconsistent}
\end{figure*}

\Cref{fiq:example-cot-inconsistent} illustrates an example of the inconsistent answers generated by ChatGPT with Chain-of-Thought \cite{kojima2022large} (CoT). It is observed that different subsets of top-8 implicit personae opinions are mentioned in the two explanations, leading to varied final answers. 

\paragraph{Human evaluation example.} \label{appdx:error-analysis}

\begin{figure*}[t!]
\centering
\includegraphics[width=1\textwidth,trim={0cm 0cm 0cm 0cm},clip]{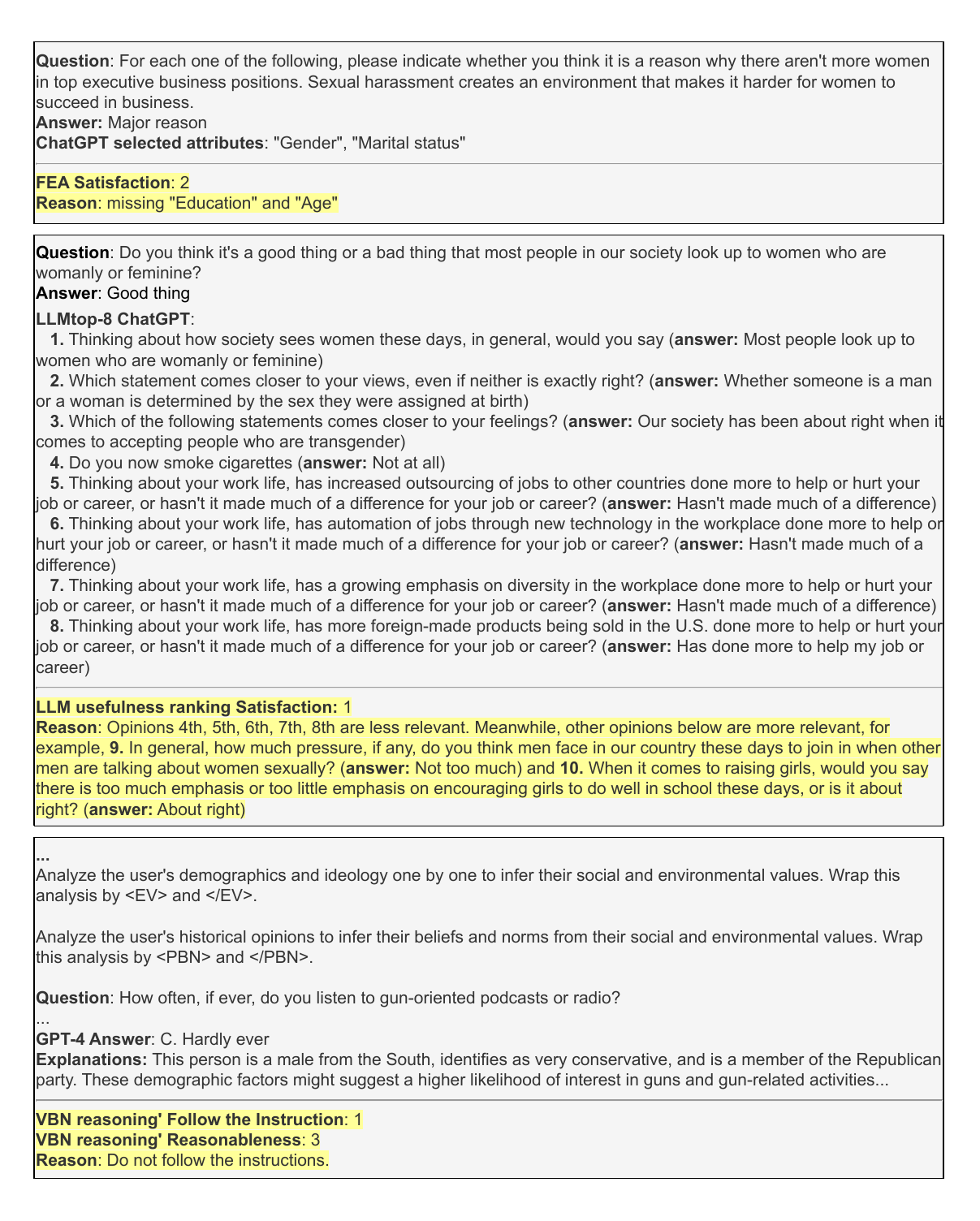}
 \caption{\small{Error analysis examples of {\model} with ChatGPT.}}
\label{fiq:error-analysis}
\end{figure*}

\Cref{fiq:error-analysis} illustrates our human evaluation example of {\model} with ChatGPT. The top-left frame is an example of FEA missing key explicit personae. The bottom one is an instance demonstrating the error of the LLMtop-$K$ algorithm including less relevant opinions. The top-right rectangular is an example from GPT-4, showing that it does not follow human instructions to predict opinion via chain-of-opinion reasoning.

\end{document}